\documentclass[11pt]{article}

\usepackage[preprint]{acl}

\usepackage{times}
\usepackage{latexsym}

\usepackage[T1]{fontenc}

\usepackage[utf8]{inputenc}

\usepackage{microtype}

\usepackage{inconsolata}

\usepackage{graphicx}
\usepackage{booktabs}
\usepackage{multirow}
\usepackage{pifont}
\usepackage{makecell}
\usepackage{booktabs}
\usepackage[table]{xcolor}
\usepackage{amsmath}
\usepackage{tabularx}
\usepackage[skins]{tcolorbox}
\usepackage{enumitem}
\usepackage{makecell}

\usepackage{algorithm}
\usepackage{algorithmic}
\usepackage{amsmath}

\definecolor{algblue}{RGB}{0, 80, 160}
\definecolor{alggray}{gray}{0.35}

\title{Cognition on Graph: Navigating Massive Knowledge Space via Cognitive Cycles and Bidirectional Graph-Text Synergy}

\author{
  \textbf{Gengxian Zhou}\textnormal{\textsuperscript{1,2}} \quad
  \textbf{Jian Xu}\textnormal{\textsuperscript{3,4}} \quad
  \textbf{Zichen Tang}\textnormal{\textsuperscript{1}} \quad
  \textbf{Shiming Xiang}\textnormal{\textsuperscript{3,4}}
\\
  \textbf{Haihong E}\textnormal{\textsuperscript{1,}\thanks{Corresponding authors.}} \quad
  \textbf{Cheng-Lin Liu}\textnormal{\textsuperscript{3,4,2,}\footnotemark[1]}
\\
  \textsuperscript{1}Beijing University of Posts and Telecommunications
\\
  \textsuperscript{2}Zhongguancun Academy, Beijing
\\
  \textsuperscript{3}MAIS, Institute of Automation, Chinese Academy of Sciences
\\
  \textsuperscript{4}School of Artificial Intelligence, University of Chinese Academy of Sciences
\\
  \texttt{Correspondence:}
  \texttt{ehaihong@bupt.edu.cn} \quad
  \texttt{liucl@nlpr.ia.ac.cn}
}

\begin{document}
\maketitle
\begin{abstract}
  Retrieval-Augmented Generation (RAG) has empowered Large Language Models (LLMs) to tackle knowledge-intensive tasks.
  However, navigating global, heterogeneous knowledge bases (large-scale knowledge graphs and text corpora) for complex reasoning remains a challenge.
  Existing methods typically employ reactive, graph-driven exploration strategies, which blindly follow graph topology without adapting to the question context or evolving exploration progress, and lack deep bidirectional synergy between graph and text.
  To address these limitations, we propose CoG (Cognition on Graph), a cognitive-inspired, training-free framework for adaptive knowledge exploration.
  Drawing inspiration from human problem-solving, CoG performs a continuous \textit{plan-explore-reflect} cycle, where it proactively formulates investigation plans, performs dual-source retrieval, and dynamically reflects on progress to adjust strategies.
  Crucially, it establishes deep bidirectional synergy between structured graph and unstructured text, where entities extracted from text dynamically guide graph exploration to bridge knowledge gaps.
  Extensive experiments on seven multi-hop QA benchmarks demonstrate that CoG significantly outperforms state-of-the-art methods while achieving superior exploration efficiency.
  Our code and datasets are available at \url{https://github.com/zhougengxian/CoG}.
\end{abstract}

\section{Introduction}
While Large Language Models (LLMs) excel in various tasks, their reliance on static parametric knowledge causes factual hallucinations and opaque reasoning in knowledge-intensive scenarios~\cite{hallucination_survey, rag_reasoning_survey, wang2025hitchhiker}. 
Retrieval-Augmented Generation (RAG) mitigates this by grounding response in external knowledge sources to improve accuracy and traceability \cite{GraphRAG_survey, knowledge_injection_survey}.

\begin{figure}[t]
  \centering
  \includegraphics[width=\columnwidth]{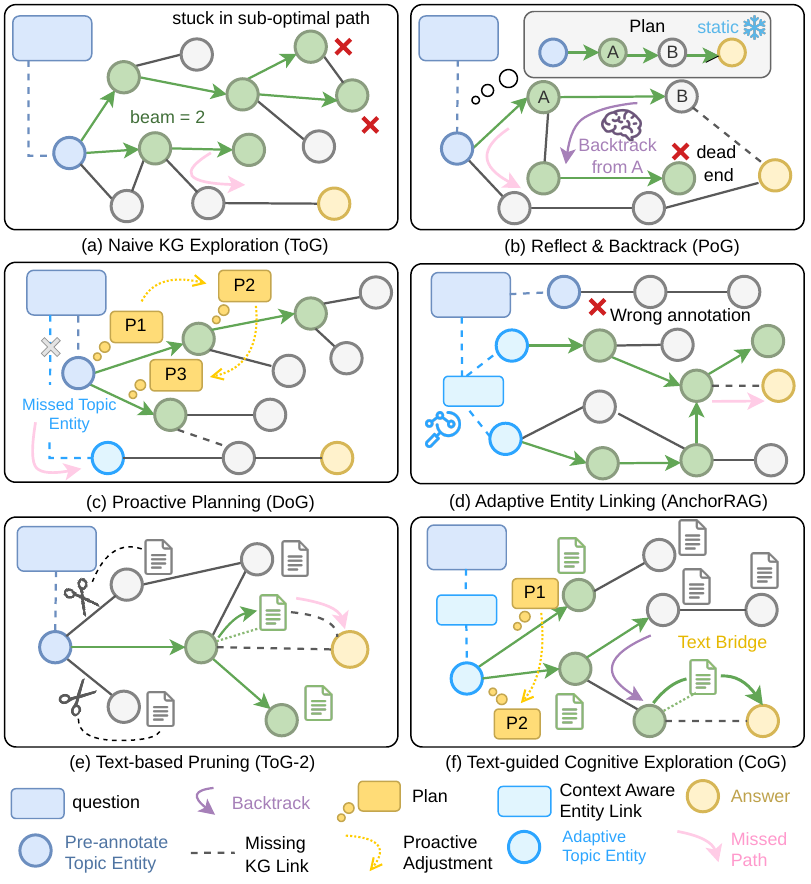}
  \vspace{-15pt}  
  \caption{Comparison of exploration strategies. CoG integrates all prior capabilities in a cognitive cycle, uniquely establishing a bidirectional synergy where text actively bridges graph gaps, enabling robust navigation in global knowledge bases.}
  \vspace{-15pt}
  \label{fig:method_comparison}
\end{figure}

Effective retrieval is the core of RAG. 
VectorRAG~\cite{RAG_work} retrieves text chunks via vector similarity, while GraphRAG \cite{HippoRAG2, GFM-RAG, HippoRAG, GraphSearch, ToG3, YoutuGraphRAG} constructs Knowledge Graphs (KGs) for structural guidance.
However, existing methods predominantly operate in simplified, local settings.
They restrict the knowledge scope to pre-selected oracle paragraphs known to contain answers, rather than raw, complete entity documents.
This masks the inherent noise of real-world retrieval, failing to assess true robustness \cite{GlobalRAG}.

When transitioning to the global setting, the core challenge escalates from simple information retrieval to complex knowledge exploration.
It requires systems to progressively construct evidence chains by synthesizing structural associations from KGs and semantic information from text corpora.
Throughout this process, the systems must effectively navigate the noise, ambiguity and distractions inherent in large-scale knowledge bases.

Existing works attempting to address this challenge, however, fall short in two critical aspects (illustrated in Figure~\ref{fig:method_comparison}).
First, exploration remains heavily \textbf{KG-dependent} with \textbf{insufficient text utilization}. 
Most methods rely solely on KG traversal for exploration \cite{ToG, PoG, DoG, ReKnoS}.
Even hybrid methods \cite{ToG2} merely use text as auxiliary context, failing to leverage entities within text to actively guide the exploration, leaving the inherent sparsity and incompleteness of KGs unaddressed.
Second, the exploration strategies are \textbf{fragile and rigid}. Relying on pre-annotated entities as rigid starting points \cite{AnchorRAG} and lacking mechanisms for backtracking \cite{PoG} or proactive adjustment \cite{DoG}, these systems are prone to getting lost in irrelevant subgraphs or stranded by linking failures or knowledge gaps.

To bridge these gaps, we draw inspiration from the cognitive process of human problem-solving.
Facing complex questions in open domains, humans do not blindly traverse information; instead, they engage in a dynamic cycle of planning, exploration, and reflection. 
They proactively formulate investigation plans, flexibly synthesize information from diverse sources, and continuously reflect on accumulated evidence to adjust their strategy.

Guided by this mechanism, we propose CoG (Cognition on Graph), a framework that instantiates this human-like reasoning process into a continuous cognitive loop comprising Planning, Exploration, Synthesis, and Reflection.
In the \textbf{Planning} phase, instead of relying on pre-annotated entities, CoG proactively analyzes the question to formulate an investigation strategy, from which it derives exploratory sub-queries and identifies relevant entities to anchor the search.
During the \textbf{Exploration} phase, CoG performs dual-source knowledge gathering. It retrieves 1-hop relevant subgraphs from the KG while simultaneously conducting hierarchical reading of entity-related documents in the text corpus, capturing both structural relations and rich contextual details.
The \textbf{Synthesis} module then aggregates evidence from both sources into a unified retrieval summary, distilling key insights that advance the reasoning process and achieving deep bidirectional synergy between structured and unstructured knowledge.
Finally, the \textbf{Reflection} module evaluates the utility of the retrieved information: for productive explorations, it updates the strategy and generates the next query based on the progress made; for unproductive ones, it diagnoses the failure using the global interaction history and adjusts the course, ensuring robust navigation through vast knowledge spaces.

Our main contributions are threefold: 
\begin{itemize}[itemsep=2pt, parsep=0pt, topsep=2pt]
\item We propose CoG, a cognitive-inspired, closed-loop, training-free GraphRAG framework. It leverages continuous cognitive cycles of plan-explore-reflect for \textbf{adaptive knowledge exploration} in noisy global knowledge bases.
\item We establish deep \textbf{bidirectional synergy} between structured and unstructured knowledge, where entities dynamically extracted from text actively guide graph exploration to effectively address KG incompleteness.
\item Extensive experiments on seven knowledge-intensive, multi-hop QA benchmarks demonstrate that CoG significantly outperforms state-of-the-art methods, showcasing superior reasoning capability and robustness in navigating large-scale knowledge bases.
\end{itemize}

\section{Related Work}
\begin{table*}[t]
  \centering
  \small
  \resizebox{\textwidth}{!}{%
  \begin{tabular}{l|c|cc|ccc|cccc|l}
  \toprule
  \multirow{2}{*}{\textbf{Method}} & \multirow{2}{*}{\textbf{Scope}} & \multicolumn{2}{c|}{\textbf{Text Corpus}} & \multicolumn{3}{c|}{\textbf{Knowledge Graph}} & \multicolumn{4}{c|}{\textbf{Key Capabilities}} & \multirow{2}{*}{\textbf{Exploration Strategy}} \\
  \cmidrule(lr){3-4} \cmidrule(lr){5-7} \cmidrule(lr){8-11}
   & & \textbf{\#Words} & \textbf{\#Ent.} & \textbf{Source} & \textbf{\#Ent.} & \textbf{\#Trip.} & \textbf{\makecell{Text\\Guide}} & \textbf{\makecell{Refl.\\\& Bk.}} & \textbf{\makecell{Pro.\\Plan.}} & \textbf{\makecell{Adapt.\\EL.}} & \\
  \midrule
  VectorRAG & Local & 1.5M & 21K & -- & -- & -- & -- & -- & -- & -- & Dense Retrieval \\
  GraphRAG Series$^{\dagger}$ & Local & 0.9M & 12K & Constr.$^{\ddagger}$ & 97K & 1.4M & -- & -- & -- & -- & Graph-Guided Retrieval \\
  \midrule
  StructGPT, KG-Agent & Global & -- & -- & Freebase & 39M & 1.9B & \ding{55} & \ding{55} & \ding{55} & \ding{55} & Interaction History \\
  ToG, ReKnoS & Global & -- & -- & Wikidata & 115M & 1.7B & \ding{55} & \ding{55} & \ding{55} & \ding{55} & Beam Search \\
  PoG & Global & -- & -- & Freebase & 39M & 1.9B & \ding{55} & \ding{51} & \ding{55} & \ding{55} & Beam Search w/ Backtrack \\
  DoG & Global & -- & -- & Freebase & 39M & 1.9B & \ding{55} & \ding{55} & \ding{51} & \ding{55} & Multi-Agent Debate \\
  AnchorRAG & Global & -- & -- & Freebase & 39M & 1.9B & \ding{55} & \ding{55} & \ding{55} & \ding{51} & Parallel Anchor Exploration \\
  KERAG & Global & -- & -- & Wikidata & 115M & 1.7B & \ding{55} & \ding{55} & \ding{55} & \ding{51} & Schema-Guided Expansion \\
  ToG-2 & Global & 5B & 7.1M & Wikidata & 115M & 1.7B & \ding{55} & \ding{55} & \ding{55} & \ding{55} & Beam Search \\
  \midrule
  \rowcolor{lightgray!30} 
  \textbf{CoG (Ours)} & \textbf{Global} & \textbf{5B} & \textbf{7.1M} & \textbf{Wikidata} & \textbf{115M} & \textbf{1.7B} & \ding{51} & \ding{51} & \ding{51} & \ding{51} & \textbf{Cognitive Cycle} \\
  \bottomrule
  \end{tabular}%
  }
  \caption{Comparison of retrieval and exploration methods. We report knowledge base scales as total \textbf{Words} and \textbf{Ent}ity documents for Text, versus \textbf{Ent}ities and \textbf{Trip}les for KG (Units: K=Thousand, M=Million, B=Billion). \textbf{CoG} uniquely operates on a massive scale while integrating all key exploration capabilities: \textbf{Text Guide} (Text entity guides graph exploration), \textbf{Refl. \& Bk.} (Reflection \& Backtrack), \textbf{Pro. Plan.} (Proactive planning), and \textbf{Adapt. EL.} (Adaptive entity linking). $\dagger$: Includes HippoRAG, HippoRAG2, GFM-RAG, et al. $\ddagger$: Graphs constructed from local text corpora. \ding{51}: Present, \ding{55}: Absent, ``--'': not applicable.}
  \vspace{-10pt}
  \label{tab:comparison}
\end{table*}

RAG enhances LLMs by retrieving external knowledge.
Beyond VectorRAG \cite{RAG_work}, GraphRAG methods construct structured indices like knowledge graphs \cite{HippoRAG, LightRAG, HippoRAG2, KG2RAG}, hierarchical communities \cite{GraphRAG_MS, HiRAG}, or document graphs \cite{KGP, KG-Retriever, GraphReader} to guide retrieval. 
Recent advancements employ reinforcement learning \cite{Graph-R1, GraphRAG-R1} or hybrid dense retrieval \cite{MoR} over graphs. 
However, these works primarily operate in local settings, restricting the knowledge base to highly relevant oracle documents or domain-specific KGs.
This simplifies the exploration challenge, failing to assess robustness in noisy global environments. 
In contrast, CoG addresses the realistic and massive scale of global knowledge bases (Wikidata and Wikipedia), navigating an exponentially larger and noisier search space.

For global knowledge exploration, prior works focus on iterative KG reasoning. 
Path-based methods like ToG~\cite{ToG}, StructGPT~\cite{StructGPT}, and KG-Agent~\cite{KG-Agent} perform prompt-driven exploration, while ReKnoS~\cite{ReKnoS} and KERAG~\cite{KERAG} improve coverage via super-relations or schemas. 
Strategy-based methods introduce specific heuristics, such as backtracking in PoG~\cite{PoG} or multi-agent debate and parallelism in DoG~\cite{DoG} and AnchorRAG~\cite{AnchorRAG}.
However, these approaches treat advanced reasoning behaviors as fragmented add-ons to fixed pipelines.
Meanwhile, learning-based methods~\cite{GRAIL, KG-R1, KGFR} train specialized navigation modules, introducing high training costs and instability while lacking explicit interpretability.
Distinguishing itself from these rigid pipelines and costly training, CoG introduces a training-free, closed-loop cognitive cycle. It natively unifies ``plan-explore-reflect'', enabling proactive planning and adaptive error-correction to emerge autonomously.

Despite advanced reasoning, KG-centric methods remain vulnerable to KG sparsity and incompleteness.
To address this, hybrid methods integrate KGs with text corpora~\cite{HybGRAG, HybridRAG, HydraRAG}.
CoK~\cite{CoK} verifies reasoning across multi-source bases, while ToG-2~\cite{ToG2} uses graph entities to anchor document retrieval.
Other approaches utilize text merely for relevance scoring~\cite{MoR} or rely on pre-constructed hyperedges~\cite{Graph-R1}.
However, these methods treat text as passive context or require rigid offline constructions, lacking dynamic structural guidance.
In contrast, CoG introduces a deep bidirectional synergy where unstructured text actively guides graph traversal. 
By extracting entities from text to dynamically bridge KG structural gaps, it achieves robust exploration in noisy global spaces.

Table~\ref{tab:comparison} provides a systematic comparison of these methods across key capabilities.

\section{Method}
\subsection{Task Definition}
\begin{figure*}[t]
  \centering
  \includegraphics[width=\textwidth]{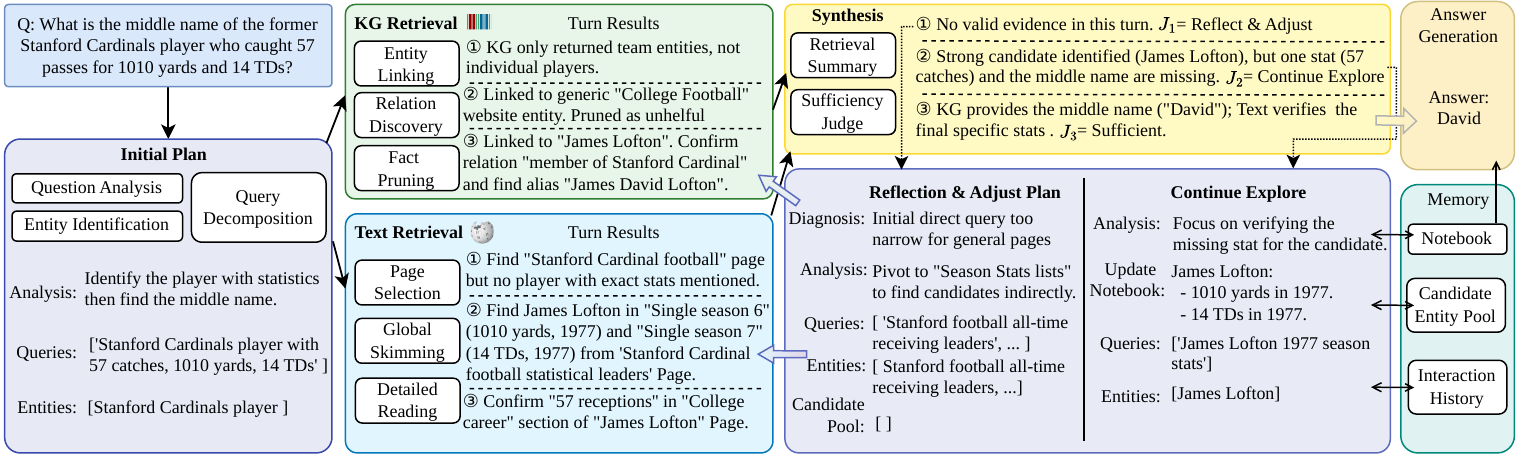} 
\caption{Overview of the CoG framework. It employs a continuous cognitive cycle of plan-explore-reflect to adaptively navigate structured and unstructured knowledge for complex QA.}
  \vspace{-10pt}
  \label{fig:framework}
\end{figure*}

We focus on the task of multi-hop Question Answering (QA) over a global, hybrid knowledge base. Given a complex question $q$, the system must construct an evidence chain by iteratively retrieving and reasoning over heterogeneous knowledge sources to generate the final answer $a$.

The knowledge base $\mathcal{K} = \{\mathcal{G}, \mathcal{D}\}$ comprises two complementary components:

\textbf{Structured Knowledge Graph} $\mathcal{G} = (\mathcal{E}, \mathcal{R}, \mathcal{T})$, where $\mathcal{E}$ and $\mathcal{R}$ denote the sets of entities and relations. The KG provides structured information in two forms: 1) relational fact triples $\mathcal{T} = \{(h, r, t) \mid h, t \in \mathcal{E}, r \in \mathcal{R}\}$ that connect pairs of entities, and 2) entity attributes (e.g., date of birth, gender) that describe the properties of individual entities.

\textbf{Unstructured Text Corpus} $\mathcal{D}$, a large-scale collection of documents. Each document $d \in \mathcal{D}$ corresponds to an entity and is hierarchically structured, containing a summary paragraph, a table of contents, and the full text organized into sections. It provides rich, fine-grained semantic descriptions that complement the factual statements in $\mathcal{G}$.

\subsection{CoG Framework}
Figure~\ref{fig:framework} illustrates the overall architecture of CoG. Given a question $q$, CoG iteratively executes a cognitive cycle until sufficient evidence is gathered to generate an answer. Each cycle consists of four tightly integrated phases: (1) Planning formulates an investigation strategy and generates a sub-query; (2) Exploration retrieves evidence from both KG and text; (3) Synthesis aggregates and distills the retrieved information; (4) Reflection evaluates progress and decides the next action.

Throughout this process, CoG maintains three long-term memory components to enable coherent multi-turn reasoning: a \textit{Notebook} $\mathcal{N}$ to accumulate verified facts, a \textit{Candidate Entity Pool} $\mathcal{P}$ to manage promising but deferred exploration leads, and an \textit{Interaction History} $\mathcal{H}$ to enable global reflection and strategic adjustment.

\subsubsection{Planning}
\label{sec:planning}
The Planning Module proactively analyzes the question and formulates an initial exploration strategy. Unlike methods that rely on pre-annotated entities, CoG dynamically derives a strategic plan directly from the question $q$. This process comprises three key components:

\textbf{Question Analysis.} It analyzes the question $q$ to identify information needs and reasoning dependencies, determining what knowledge must be gathered first. This analysis $A_0$ serves as a high-level guide for subsequent actions.

\textbf{Query Decomposition.} Based on the analysis, the question is decomposed into a set of focused sub-queries $Q_0 = \{q_1^{(0)}, \dots, q_B^{(0)}\}$, each targeting a specific information need that can be explored independently as the immediate next step.

\textbf{Entity Identification.} It identifies a list of anchor entities $E_0 = \{e_1^{(0)}, \dots, e_B^{(0)}\}$ corresponding one-to-one with $Q_0$. Each $e_i^{(0)}$ serves as the core anchor point for initiating the exploration of query $q_i^{(0)}$ in both the KG and text corpus.

Formally, the initial plan $P_0$ is generated by:
\begin{equation}
    P_0 = (A_0, Q_0, E_0) = \text{Prompt}_{\text{init}}(q, \text{LLM}).
\end{equation}

The prompt is provided in Appendix~\ref{sec:prompts}, Table~\ref{tab:prompt_init_plan}.

\subsubsection{Exploration}
\label{sec:exploration}

Given a sub-query $q_i^{(t)}$ and its anchor entity $e_i^{(t)}$ from the Planning phase, the Exploration module performs \textbf{dual-source retrieval} to gather complementary evidence from both the structured KG $\mathcal{G}$ and the unstructured text corpus $\mathcal{D}$.

\paragraph{KG Exploration.} 
To navigate massive global KGs with inherent ambiguity and noise, we design a three-stage pipeline that progressively narrows the search scope from entity disambiguation to relation filtering and fine-grained fact pruning.

\textbf{(1) Entity Linking.} We design a context-aware coarse-to-fine recall strategy. First, we retrieve top-$k$ candidate entities $\mathcal{C}$ via vector similarity between the query context $C_i^{(t)} = (q, A_t, q_i^{(t)}, e_i^{(t)})$ and entity profiles (aliases, description and neighbors). Then, the LLM performs precise disambiguation by evaluating each candidate's profile against the query context to determine the optimal match:
\begin{align}
  \mathcal{C} &= \text{VectorRecall}(C_i^{(t)}, \mathcal{E}, k), \nonumber \\
  e^* &= \text{Prompt}_{\text{link}}(C_i^{(t)}, \mathcal{C}, \text{LLM}).
\end{align}

\textbf{(2) Relation Discovery.} First, we retrieve all relations $R_e$ connected to the linked entity $e^*$. Then, the LLM selects a subset $R_{\text{sel}} \subset R_e$ relevant to the query context $C_i^{(t)}$ by considering relation labels, connectivity and potential to lead toward the answer, thereby pruning the search space at the schema level:
\begin{align}
  R_{e^*} &= \text{GetRelations}(e^*, \mathcal{G}), \nonumber \\
  R_{\text{sel}} &= \text{Prompt}_{\text{rel}}(C_i^{(t)}, e^*, R_{e^*}, \text{LLM}).
\end{align}

\textbf{(3) Fact Pruning.} We retrieve all facts $F_{raw}$ (triples and attributes) associated with the selected relations $R_{\text{sel}}$. The LLM performs fine-grained pruning to retain only facts that either directly answer the query $q_i^{(t)}$ or serve as promising stepping stones for further exploration, filtering out noise while preserving critical evidence:
\begin{align}
  F_{raw} &= \{(e^*,r,t) \in \mathcal{T} \mid r \in R_{\text{sel}}, t \in \mathcal{E}\} \nonumber \\
          &\quad \cup \{(h,r,e^*) \in \mathcal{T} \mid r \in R_{\text{sel}}, h \in \mathcal{E}\}, \nonumber \\
  F_{i}^{KG} &= \text{Prompt}_{\text{fact}}(C_i^{(t)}, F_{raw}, \text{LLM}).
\end{align}

\paragraph{Text Exploration.} We propose a hierarchical reading strategy that mimics human reading behavior, narrowing the focus from document retrieval to global skimming and fine-grained section analysis, balancing retrieval efficiency with depth.

\textbf{(1) Adaptive Page Selection.} We employ an adaptive mechanism to retrieve the optimal entity document $d_{i}$ for $e_i^{(t)}$. When encountering disambiguation or missing pages, the LLM progressively refines the search query by analyzing query context $C_i^{(t)}$ and Wikipedia API feedback, ensuring robust page localization despite real-world noise.

\textbf{(2) Global Skimming.} The LLM skims the high-level structure of the retrieved page $d_i$ ( summary, table of contents and infobox) to assess the page's relevance to the query context $C_i^{(t)}$ and extract useful evidence $I_{skim}$. If deeper investigation is needed, it selects a focused subset of promising sections $S_{sel}$ likely to contain missing details:
\begin{equation}
  S_{sel}, I_{skim} = \text{Prompt}_{\text{skim}}(C_i^{(t)}, d_i, \text{LLM}).
\end{equation}

\textbf{(3) Detailed Reading.} For each selected section $s \in S_{sel}$, we retrieve its full content,  split it into chunks, and identify the top-$k$ relevant chunks via dense retrieval. Concurrently, tables within the section are extracted and serialized. Finally, the LLM performs joint analysis of text chunks and tables to synthesize fine-grained evidence $I_{detail}^{s}$:
\begin{equation}
  I_{detail}^{(s)} = \text{Prompt}_{\text{detail}}(C_i^{(t)}, s, \text{LLM}).
\end{equation}

Combining skimmed and detailed extractions, the final text evidence for entity $e_i^{(t)}$ is $F_i^{text} = I_{skim} \cup \{I_{detail}^{(s)} \mid s \in S_{sel}\}$.

\paragraph{Dual-Source Evidence.} The final retrieved evidence for sub-query $q_i^{(t)}$ is the aggregation of dual-source knowledge: $K_i^{(t)} = F_{i}^{KG} \cup F^{text}_i$.

Prompts for KG (Tables~\ref{tab:prompt_entity_link}--\ref{tab:prompt_fact_prune}) and text  (Tables~\ref{tab:prompt_disambiguate}--\ref{tab:prompt_detail}) exploration are provided in Appendix~\ref{sec:prompts}.

\subsubsection{Synthesis \& Reflection}
\label{sec:synthesis_reflection}

The \textbf{Synthesis} module aggregates dual-source evidence and evaluates progress. 
We design structured formats to present KG and text retrieval $K_t = \{K_1^{(t)}, \dots, K_B^{(t)}\}$.
The LLM first distills key facts and promising leads into a unified retrieval summary $M_t$, achieving deep synergy between structured KG and unstructured text.
Simultaneously, it makes a judgment $J_t$ based on the notebook $\mathcal{N}$ and the current plan $P_t=(A_t, Q_t, E_t)$ to determine whether to generate the final answer, continue exploration, or adjust the strategy:
\begin{equation}
  J_t, M_t = \text{Prompt}_{\text{syn}}(q, P_t, K_t, \mathcal{N}, \text{LLM}).
\end{equation}

Depending on $J_t$, the \textbf{Reflection} module dynamically directs exploration through two pathways:

\textbf{(1) Continue Exploration.} When progress is made but evidence remains incomplete, it refines the analysis to reflect new findings, formulates a new plan $P_{t+1}$ targeting remaining knowledge gaps. Simultaneously, it updates two long-term memory structures: Notebook $\mathcal{N}$, which accumulates verified key facts directly relevant to $q$, and Candidate Entity Pool $\mathcal{P}$, which stores promising but deferred entities due to exploration width constraints:
\begin{equation}
  P_{t+1}, \mathcal{N}, \mathcal{P} = \text{Prompt}_{\text{cont}}(q, P_t, M_t, \mathcal{N}, \mathcal{P}, \text{LLM}).
\end{equation}

\textbf{(2) Strategy Adjustment.} When facing dead ends, it performs critical reflection by analyzing the global interaction history $\mathcal{H}$ to diagnose strategic errors. Then it pivots the strategy by backtracking to deferred entities in $\mathcal{P}$ or reformulating queries from entirely new angles based on the $q$ and current progress, thereby ensuring resilient exploration:
\begin{equation}
  P_{t+1}, \mathcal{P} = \text{Prompt}_{\text{recov}}(q, P_t, M_t, \mathcal{H}, \mathcal{N}, \mathcal{P}, \text{LLM}).
\end{equation}

\textbf{Interaction History Update.} After each turn, the system appends the current retrieval summary $M_t$ and plan $P_t$ to the interaction history: $\mathcal{H} \leftarrow \mathcal{H} \cup \{(P_t, M_t)\}$. 
This cumulative memory enables global retrospection, allowing the system to avoid repeating failed strategies and to make globally informed decisions across multiple turns.

Prompts for synthesis (Table~\ref{tab:prompt_synthesis}) and reflection (Tables~\ref{tab:prompt_cont_explore}, \ref{tab:prompt_recover}) are provided in Appendix~\ref{sec:prompts}.

\subsubsection{Overall Workflow \& Answer Generation}
\label{sec:exploration_answer_generation}

CoG executes the aforementioned cognitive cycle iteratively, until the Reflection module deems the evidence sufficient or a maximum turn limit is reached. The complete algorithmic workflow is detailed in Algorithm~\ref{alg:cog_framework} in Appendix~\ref{sec:cog_algorithm}.

Upon sufficiency, the system synthesizes accumulated facts in Notebook $\mathcal{N}$ and final retrieval result $M_t$. The LLM constructs a comprehensive reasoning chain connecting these facts to directly address the question, producing the final answer $a$.

\section{Experiments}
\subsection{Experimental Setup}

\paragraph{Datasets.} We evaluate on seven knowledge-intensive multi-hop QA benchmarks:
(1) KG-based QA: KGQAGen~\cite{KGQAGen}, CWQ~\cite{CWQ}, QALD10-en~\cite{QALD10}, and WebQSP~\cite{WebQSP}, which require reasoning over structured facts for entity-centric answers.
(2) Text-based QA: 2WikiMQA~\cite{2WikiMQA}, AdvHotpotQA~\cite{AdvHotpotQA}, and MusiQue~\cite{MusiQue}, which demand deep semantic understanding and multi-document synthesis for span extraction and free-form answer generation.
Detailed descriptions and statistics are provided in Appendix~\ref{sec:evaluation_dataset}.
We use Exact Match (EM) as the primary evaluation metric following previous work~\cite{ToG2}, and additionally report token-level F1 and reference-based LLM-as-judge results in Appendix~\ref{sec:additional_metrics}.
For repeated runs, we report the median of three independent runs to ensure robustness.

\paragraph{Baselines.} We compare CoG against representative methods across four categories:
(1) LLM-only: Direct few-shot prompting, Chain-of-Thought~\cite{CoT}, and Self-Consistency~\cite{CoT-SC}, relying solely on internal parametric knowledge;
(2) Text-based RAG: VectorRAG (top-$3$ chunk dense retrieval), ReAct~\cite{ReAct} and IRCoT~\cite{IRCoT} for iterative text exploration, and Search-o1~\cite{Search-o1} for on-demand retrieval during reasoning;
(3) KG-based RAG: ToG~\cite{ToG} for iterative KG traversal, PoG~\cite{PoG} with backtracking, and DoG~\cite{DoG} using debate-driven query evolution;
(4) Hybrid RAG: ToG-2~\cite{ToG2}, the current state-of-the-art framework combining KG traversal with text retrieval under the same global knowledge setting as CoG.

\paragraph{Implementation Details.}
Following previous works~\cite{ToG2, CoK}, we adopt the global knowledge setting, utilizing the full Wikidata and Wikipedia as structured and unstructured knowledge bases, respectively (details in Appendix~\ref{sec:knowledge_bases}). This setting presents a realistic and challenging environment compared to simplified local settings.
We employ Qwen3-32B (non-thinking mode) as the backbone LLM for CoG and baselines to ensure fair comparison.
For retrieval, we utilize bge-m3 \cite{bge-m3} as the embedding model for text chunk retrieval, and Qwen3-Embedding-4B \cite{Qwen3-embed} for entity linking in the KG.
We set the maximum exploration depth to 4 and width (sub-queries per turn) to 5. For entity linking, we retrieve the top-20 candidate entities. In text exploration, we extract up to 3 relevant chunks per section.

\subsection{Main Results}
\label{sec:main_results}

\begin{table*}[!t]
  \centering
  \small
  \renewcommand{\arraystretch}{1.05}
  \resizebox{\textwidth}{!}{
  \begin{tabular}{llccccccc}
      \toprule
      \multirow{2}{*}{\textbf{Baseline Type}} & \multirow{2}{*}{\textbf{Method}} & \multicolumn{4}{c}{\textbf{Multi-hop KG-based QA}} & \multicolumn{3}{c}{\textbf{Multi-hop text-based QA}} \\
      \cmidrule(lr){3-6} \cmidrule(lr){7-9}
      & & KGQAGen & CWQ & QALD10-en & WebQSP & 2WikiMQA & AdvHotpotQA & MusiQue \\
      \midrule
      \multirow{3}{*}{LLM-only} 
      & Direct & 36.70 & 32.42 & 47.45 & 72.45 & 47.20 & 20.13 & 8.40 \\
      & CoT & 38.28 & 33.59 & 45.35 & 70.84 & 47.40 & 23.70 & 9.60 \\
      & CoT-SC & 40.50 & 33.40 & 46.55 & 71.40 & 49.60 & 24.03 & 10.60 \\
      \midrule
      \multirow{4}{*}{Text-based RAG} 
      & Vector RAG & 48.66 & 30.76 & 39.04 & 70.49 & 52.60 & 32.14 & 12.40 \\
      & ReAct & 50.79 & 34.18 & 41.44 & 76.15 & 63.20 & 41.88 & 17.00 \\
      & IRCoT & 53.75 & 33.30 & 41.14 & 71.40 & 71.60 & 47.08 & 20.60 \\
      & Search-o1 & 46.34 & 36.72 & 43.54 & 72.03 & 44.60 & 40.26 & 19.40 \\
      \midrule
      \multirow{3}{*}{KG-based RAG} 
      & ToG & 46.90 & 28.42 & 48.65 & 73.28 & 50.40 & 18.83 & 9.20 \\
      & PoG & 50.93 & 27.15 & 48.95 & 78.53 & 73.20 & 15.26 & 14.00 \\
      & DoG & 54.77 & 33.11 & \textbf{55.86} & \textbf{82.38} & 61.20 & 20.13 & 11.40 \\
      \midrule
      Hybrid RAG 
      & ToG-$2$ & 55.61 & 41.41 & 53.45 & 81.89 & 73.00 & 35.71 & 16.20 \\
      \rowcolor{blue!10}
      \multirow{1}{*}{Proposed}
      & CoG & \textbf{74.70} & \textbf{42.87} & 55.26 & 81.75 & \textbf{85.20} & \textbf{51.95} & \textbf{27.80}\\
      \bottomrule
  \end{tabular}
  }
  \caption{Main results (Exact Match, \%) of different methods on seven multi-hop QA datasets. All methods use Qwen3-32B (non-thinking mode) as the backbone LLM. The best results are highlighted in \textbf{bold}.}
  \label{tab:main_results}
\end{table*}

As shown in Table~\ref{tab:main_results}, CoG substantially outperforms baselines across most benchmarks, with advantages particularly pronounced on complex tasks. 
On the challenging text-heavy MusiQue, CoG (27.8\%) significantly surpasses the strongest text agent IRCoT (20.6\%). 
Similarly, it achieves a 34.3\% relative gain over ToG-2 on the high-quality KG-based KGQAGen. 
Meanwhile, CoG maintains comparable performance on simpler WebQSP (1-2 hops). 
These results validate our cognitive framework's effectiveness in unifying structured and unstructured knowledge exploration.
Additional F1 and LLM-as-judge results in Appendix~\ref{sec:additional_metrics} confirm that these gains persist under both lexical-overlap and semantic evaluation.

\paragraph{Analysis of Baselines} VectorRAG struggles with KG-based tasks, as its flat retrieval lacks structural guidance.
Agentic text baselines (ReAct, IRCoT, Search-o1) and strong KG agents (PoG, DoG) improve over basic RAG but exhibit single-source limitations: text agents falter on KG-heavy tasks, while DoG struggles on text-heavy ones.
ToG-2 effectively combines KG and text to outperform single-source baselines, but its reactive exploration lacks long-term planning and bidirectional synergy, faltering on complex tasks.
CoG consistently outperforms single-source and hybrid baselines, proving that our Graph-Text Synergy effectively overcomes pure graph limitations and text retrieval noise, yielding benefits beyond a generic agentic loop.

\subsection{Ablation Study}
\label{sec:ablation_study}

\begin{table}[t]
  \centering
  \small
  \setlength{\tabcolsep}{2.5pt} 
  \renewcommand{\arraystretch}{1.1}
  \resizebox{\columnwidth}{!}{
  \begin{tabular}{lcccccc|c}
      \toprule
      \textbf{Variant} & \textbf{KGQA} & \textbf{CWQ} & \textbf{QALD} & \textbf{2Wiki} & \textbf{AdvHot} & \textbf{MusiQ} & \textbf{Avg.} \\
      \midrule
      \textbf{CoG (Full)} & 74.70 & \textbf{42.87} & \textbf{55.26} & \textbf{85.20} & \textbf{51.95} & \textbf{27.80} & \textbf{56.30} \\
      \midrule
      \rowcolor{gray!12}
      \multicolumn{8}{c}{\textit{Impact of Knowledge Sources}} \\
      w/o Text & \textbf{75.90} & 39.06 & 51.65 & 76.20 & 27.27 & 20.20 & 48.38 \\
      w/o KG & 56.44 & 40.92 & 48.65 & 79.80 & 49.68 & 20.80 & 49.38 \\
      \midrule
      \rowcolor{gray!12}
      \multicolumn{8}{c}{\textit{Impact of Bidirectional Synergy}} \\
      w/o T-guide & 74.14 & 42.58 & 52.85 & 76.20 & 39.61 & 22.80 & 51.36 \\
      w/o Synthesis & 73.12 & 41.11 & 50.15 & 83.60 & 43.83 & 24.20 & 52.67 \\
      \midrule
      \rowcolor{gray!12}
      \multicolumn{8}{c}{\textit{Impact of Cognitive Strategy}} \\
      w/o Reflect & 72.57 & 42.38 & 52.85 & 85.00 & 48.05 & 25.80 & 54.44 \\
      w/o Plan & 78.78 & 43.85 & 50.75 & 81.00 & 47.73 & 25.40 & 54.59 \\
      w/o Adap.EL & 70.81 & 44.14 & 53.15 & 86.00 & 47.73 & 23.40 & 54.21 \\
      \bottomrule
  \end{tabular}
  }
  \caption{Ablation results (EM, \%) of CoG across three groups: (1) \textbf{Knowledge Sources}: removing either KG (\textit{w/o KG}) or Text (\textit{w/o Text}); (2) \textbf{Bidirectional Synergy}: disabling text-guided KG exploration (\textit{w/o T-guide}), or removing summary synthesis module (\textit{w/o Synthesis}); (3) \textbf{Cognitive Strategy}: excluding reflection-based adjustment (\textit{w/o Reflect}), proactive planning (\textit{w/o Plan}), or adaptive entity linking (\textit{w/o Adap.EL}).}
  \vspace{-10pt}
  \label{tab:ablation}
\end{table}

To dissect CoG's components, we conduct ablation studies across three dimensions (Table~\ref{tab:ablation}).

\paragraph{Knowledge Sources.}
Removing either source causes significant drops ($-6.92\%$ for \textit{w/o KG}, $-7.92\%$ for \textit{w/o Text}). The severe degradation of \textit{w/o Text} on text-heavy datasets (AdvHotpotQA, MusiQue) confirms that while KG provides structural guidance, rich text semantics are indispensable for bridging knowledge gaps.

\paragraph{Bidirectional Synergy.}
Simply accessing both sources is insufficient.
Disabling text-to-KG feedback (\textit{w/o T-guide}) drops performance by 4.94\%, indicating that entities discovered from text serve as effective new anchors for bridging disconnected KG paths. 
Furthermore, removing the synthesis module (\textit{w/o Synthesis}) degrades performance, indicating that explicitly distilling heterogeneous evidence is crucial for filtering noise and informing subsequent reasoning.

\paragraph{Cognitive Strategy.}
Cognitive mechanisms ensure robust exploration.
Removing reflection, planning, or adaptive entity linking all degrades average performance.
These results suggest that CoG benefits not only from access to KG and text, but also from its closed-loop control: reflection helps recover from noisy paths, planning guides goal-oriented search, and adaptive entity linking provides reliable anchors under ambiguous mentions.

\subsection{In-depth Analysis}
\label{sec:in-depth_analysis}

\paragraph{Generalizability and Cost-Effectiveness.}
Table~\ref{tab:llm_generalization} demonstrates CoG's effectiveness across diverse LLMs (detailed in Appendix~\ref{sec:appendix_llm_analysis}). 
Notably, CoG unlocks the reasoning potential of smaller models, yielding a massive +113.2\% gain on Qwen3-8B.
Remarkably, CoG-8B (51.0\%) outperforms all strong 32B baselines (e.g., IRCoT 44.6\%, ToG-2 45.9\%). 
This confirms that CoG's superiority stems from its robust cognitive architecture rather than mere parameter scale, effectively mitigating error propagation on smaller LLMs while offering a highly cost-effective solution. 

\begin{table}[h]
  \centering
  \small
  \resizebox{\columnwidth}{!}{
  \begin{tabular}{l|c|cc|c}
      \toprule
      \textbf{Backbone} & \textbf{Method} & \textbf{KG Avg.} & \textbf{Text Avg.} & \textbf{Overall Avg.} \\
      \midrule
      \multirow{2}{*}{Qwen3-32B} & Direct & 41.48 & 25.24 & 33.36 \\
      & CoG & \textbf{59.31} & \textbf{54.98} & \textbf{57.15} \scriptsize{\textcolor{teal}{(+71.3\%)}} \\
      \midrule
      \multirow{2}{*}{Qwen3-8B} & Direct & 31.09 & 16.73 & 23.91 \\
      & CoG & \textbf{53.39} & \textbf{48.55} & \textbf{50.97} \scriptsize{\textcolor{teal}{(+113.2\%)}} \\
      \midrule
      \multirow{2}{*}{gpt-oss-120b} & Direct & 28.33 & 15.88 & 22.10 \\
      & CoG & \textbf{42.87} & \textbf{40.28} & \textbf{41.58} \scriptsize{\textcolor{teal}{(+88.1\%)}} \\
      \midrule
      \multirow{2}{*}{DeepSeek-V3.2} & Direct & 47.91 & 28.06 & 37.99 \\
      & CoG & \textbf{64.79} & \textbf{60.50} & \textbf{62.64} \scriptsize{\textcolor{teal}{(+64.9\%)}} \\
      \midrule
      \multirow{2}{*}{GPT-5.1} & Direct & 45.67 & 38.67 & 42.17 \\
      & CoG & \textbf{56.00} & \textbf{55.33} & \textbf{55.67} \scriptsize{\textcolor{teal}{(+32.0\%)}} \\
      \bottomrule
  \end{tabular}
  }
  \caption{Performance comparison (EM, \%) of CoG versus Direct Prompting across different backbone LLMs. Proprietary models are evaluated on a sampled subset (100 samples per dataset) due to cost constraints.}
  \label{tab:llm_generalization}
\end{table}

\paragraph{Impact of Exploration Depth.} Figure~\ref{fig:depth_analysis} illustrates the performance trends across varying depths (see Appendix~\ref{sec:appendix_depth_analysis} for full results). 
On KG tasks, CoG consistently outperforms VectorRAG across all depths, confirming the superiority of structured cognitive exploration over flat retrieval.
Similarly, text tasks show steady gains as depth increases, indicating that deeper exploration effectively gathers scattered semantic evidence.
Notably, the optimal depth correlates strongly with dataset difficulty. 
Simpler datasets (WebQSP, 2WikiMQA) achieve optimal performance at shallower depths and then stabilize, whereas challenging ones (KGQAGen, AdvHotpotQA) continue to benefit from deeper exploration.
It demonstrates CoG's adaptability to varying task complexity.

\begin{figure}[t]
  \centering
  \includegraphics[width=0.9\columnwidth]{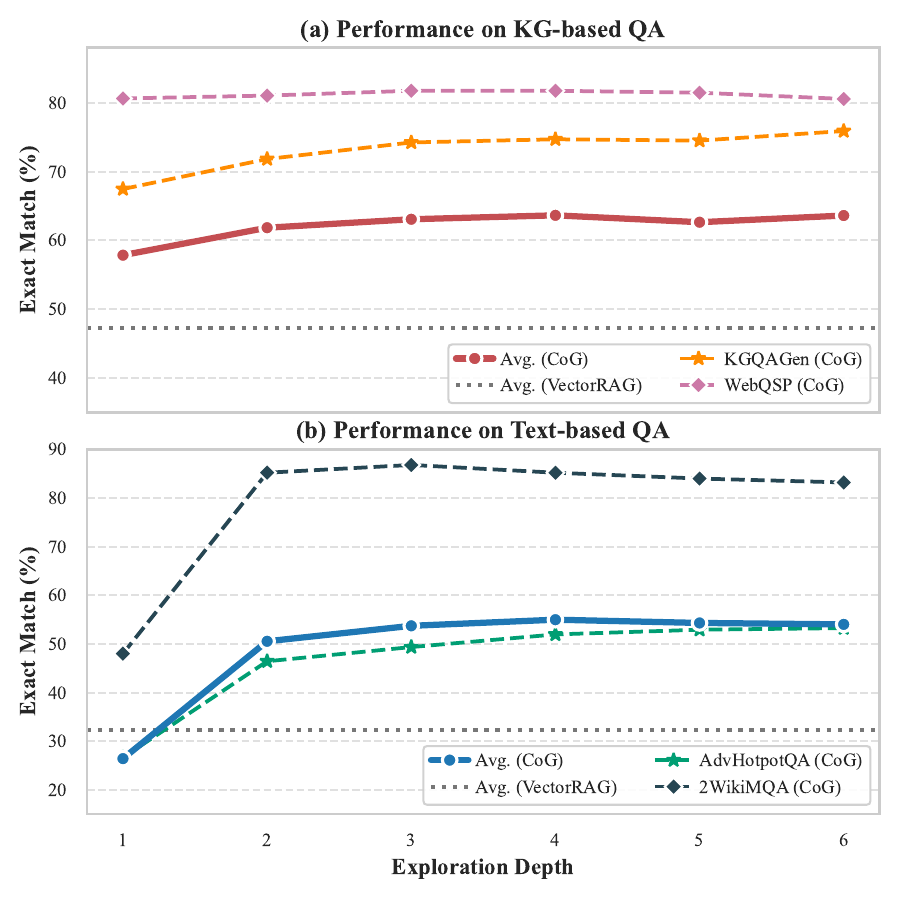}
  \caption{Impact of exploration depth on KG-based and Text-based tasks and representative datasets.}
  \vspace{-10pt}
  \label{fig:depth_analysis}
\end{figure}

\paragraph{Exploration Efficiency.}
We evaluate efficiency across four dimensions: accessed documents, token consumption, latency, and wall-clock time per retrieval round (detailed breakdown in Appendix~\ref{sec:appendix_efficiency}).
Despite multiple LLM calls per turn, CoG's goal-oriented planning achieves remarkable precision, accessing $12.4\times$ fewer documents than ToG-2's exhaustive beam search (Table~\ref{tab:doc_efficiency}).
Regarding absolute cost, CoG exhibits complexity adaptability: on simpler datasets (e.g., WebQSP), it maintains low token usage ($\sim$15k) comparable to text agents like IRCoT ($\sim$22k) while achieving superior accuracy. On complex tasks (e.g., MusiQue), costs increase to support deep reasoning, trading computation for capability where efficient methods fail.
Crucially, this overhead can be mitigated by using smaller models, as CoG-8B already outperforms 32B baselines (Appendix~\ref{sec:appendix_llm_analysis}), improving the accuracy-cost trade-off for deployment.

\begin{table}[!t]
  \centering
  \small
  \setlength{\tabcolsep}{3pt}
  \resizebox{\columnwidth}{!}{
  \begin{tabular}{l|cccc|ccc|c}
      \toprule
      \textbf{Method} & \textbf{KGQA} & \textbf{CWQ} & \textbf{QALD} & \textbf{WebQ} & \textbf{2Wiki} & \textbf{Hotpot} & \textbf{MusiQ} & \textbf{Avg.} \\
      \midrule
      ToG-2 & 37.1 & 40.4 & 31.2 & 22.9 & 17.7 & 36.1 & 42.2 & 32.5 \\
      \textbf{CoG} & \textbf{3.1} & \textbf{3.1} & \textbf{2.3} & \textbf{1.5} & \textbf{2.4} & \textbf{2.5} & \textbf{3.5} & \textbf{2.6} \\
      \midrule
      \textit{Gain} & $\times$11.9 & $\times$13.2 & $\times$13.3 & $\times$15.5 & $\times$7.3 & $\times$14.8 & $\times$12.1 & \textbf{$\times$12.4} \\
      \bottomrule
  \end{tabular}
  }
  \caption{Comparison of exploration efficiency (Average number of unique documents accessed per question).}
  \vspace{-10pt}
  \label{tab:doc_efficiency}
\end{table}

\paragraph{Hyperparameter Robustness.}
We study the sensitivity of adaptive entity linking by varying the retrieved entity candidates from $k=6$ to $30$ (Appendix~\ref{sec:appendix_el_sensitivity}).
CoG remains stable across settings, with average EM varying only slightly from 59.1\% to 60.3\%.
Increasing $k$ does not degrade performance, suggesting that CoG's context-aware disambiguation can filter noisy candidates while benefiting from broader entity recall.

\paragraph{Manual Evaluation.}
We manually analyze sampled reasoning paths in Appendix~\ref{sec:manual_evaluation} to evaluate reasoning faithfulness and diagnose error causes.
For EM-correct cases, CoG yields the highest proportion of \textit{True Positives} (89.1\%) with faithful reasoning chains. 
In contrast, CoT-SC and ToG-2 suffer more severe false positives (\textit{Weak FP + FP}: 29.1\% and 12.7\% vs. CoG's 3.6\%), often reaching correct answers via hallucinated or flawed intermediate steps.
For EM-incorrect cases, 71.1\% of CoG's errors are attributable to external factors such as metric false negatives, dataset flaws, or temporal mismatch, while 28.9\% reflect internal exploration or reasoning failures. 
These findings support both the reliability of CoG's EM gains and the need for stronger exploration focus and disambiguation.

\paragraph{Qualitative Analysis.}
We provide a detailed case study in Appendix~\ref{sec:case_studies}, demonstrating how CoG's cognitive cycle and bidirectional synergy enable it to recover from dead ends and discover critical text evidence to solve complex multi-hop queries.

\section{Conclusion}
In this paper, we introduced CoG, a cognitive-inspired framework for autonomous knowledge exploration in global, hybrid knowledge bases.
Under the continuous ``plan-explore-reflect'' human cognitive cycle, CoG shifts the paradigm from passive, reactive retrieval to proactive knowledge discovery, enabling adaptive strategy adjustment and robust recovery from exploration failures.
Crucially, it establishes a deep bidirectional synergy between structured KGs and unstructured text, leveraging their complementary strengths to bridge knowledge gaps.
Extensive experiments demonstrate that CoG significantly outperforms state-of-the-art baselines on multiple multi-hop QA tasks while achieving superior exploration efficiency.

\section*{Limitations}
\label{sec:limitations}

While CoG demonstrates superior performance and exploration efficiency in global knowledge exploration, several limitations remain.

\paragraph{Latency and Cost.} The iterative cognitive loop inevitably incurs higher inference latency and token cost than passive retrieval methods. Although CoG reduces redundant document access through goal-oriented exploration and can improve the cost-accuracy trade-off when paired with smaller backbones, its multi-call workflow may still limit real-time deployment.

\paragraph{Dependency on Backbone LLM.} The framework's effectiveness still depends on the backbone LLM's instruction following and long-context reasoning abilities. Although CoG substantially improves smaller models, weaker LLMs may still struggle with complex planning, evidence synthesis, or reflection over noisy intermediate results.

\section*{Acknowledgments}

This work is supported by the Zhongguancun Academy (Grant No. 02012501), in part by the National Natural Science Foundation of China (Grant Nos. 62473271, 62506357), and the Fundamental Research Funds for the Beijing University of Posts and Telecommunications (Grant No. 2025AI4S03).

\bibliography{custom}

\appendix

\section{CoG Algorithm}
\label{sec:cog_algorithm}

We provide the pseudo-code of the CoG framework in Algorithm~\ref{alg:cog_framework}. It outlines the iterative workflow, illustrating how the Planning, Exploration, Synthesis, and Reflection modules coordinate to achieve autonomous knowledge exploration in large-scale knowledge bases.

\begin{algorithm}[t]
  \small
  \caption{The CoG Algorithm}
  \label{alg:cog_framework}
  \renewcommand{\algorithmicrequire}{\textbf{Input:}}
  \renewcommand{\algorithmicensure}{\textbf{Output:}}
  \begin{algorithmic}
  \REQUIRE Question $q$, KG $\mathcal{G}$, Text Corpus $\mathcal{D}$, Max Turns $M$
  \ENSURE Answer $a$
  \STATE \textbf{Initialize:} $\mathcal{N} \gets \emptyset$, $\mathcal{P} \gets \emptyset$, $\mathcal{H} \gets \emptyset$
  \STATE \textcolor{algblue}{// \textbf{Phase 1: Initial Planning} (§\ref{sec:planning})}
  \STATE $P_0 = (A_0, Q_0, E_0) \gets \textsc{InitPlan}(q)$
  \FOR{$t = 0$ to $M-1$}
      \STATE \textcolor{algblue}{// \textbf{Phase 2: Exploration Dual-Source} (§\ref{sec:exploration})}
      \STATE $K_t \gets \emptyset$
      \FOR{each $(q_i^{(t)}, e_i^{(t)}) \in (Q_t, E_t)$}
          \STATE $C_i^{(t)} \gets (q, A_t, q_i^{(t)}, e_i^{(t)})$ \textcolor{alggray}{// Query context}
          \STATE $F_{i}^{KG} \gets \textsc{ExploreKG}(C_i^{(t)}, \mathcal{G})$ 
          \STATE \textcolor{alggray}{// Entity Link $\rightarrow$ Relation Discovery $\rightarrow$ Fact Prune}
          \STATE $F_{i}^{text} \gets \textsc{ExploreText}(C_i^{(t)}, \mathcal{D})$ 
          \STATE \textcolor{alggray}{// Page Selection $\rightarrow$ Skimming $\rightarrow$ Detailed Reading}
          \STATE $K_t \gets K_t \cup \{F_{i}^{KG} \cup F_{i}^{text}\}$ \textcolor{alggray}{// Aggregate dual-source evidence}
      \ENDFOR
      \STATE \textcolor{algblue}{// \textbf{Phase 3: Synthesis} (§\ref{sec:synthesis_reflection})}
      \STATE $J_t, M_t \gets \textsc{Synthesize}(q, P_t, K_t, \mathcal{N})$
      \STATE \textcolor{algblue}{// \textbf{Phase 4: Reflection \& Re-Planning} (§\ref{sec:synthesis_reflection})}
      \IF{$J_t = \texttt{SUFFICIENT}$}
      \STATE \textbf{break} \textcolor{alggray}{// Proceed to Answer Generation}
      \ELSIF{$J_t = \texttt{INSUFFICIENT\_USEFUL}$}
      \STATE \textcolor{alggray}{// Pathway 1: Continue Exploration}
      \STATE $P_{t+1}, \mathcal{N}, \mathcal{P} \gets \textsc{ContinuePlan}(q, P_t, M_t, \mathcal{N}, \mathcal{P})$
      \ELSE
      \STATE \textcolor{alggray}{// Pathway 2: Strategy Adjustment}
      \STATE $P_{t+1}, \mathcal{P} \gets \textsc{AdjustPlan}(q, P_t, M_t, \mathcal{H}, \mathcal{N}, \mathcal{P})$
      \ENDIF
      \STATE $\mathcal{H} \gets \mathcal{H} \cup \{(P_t, M_t)\}$ \textcolor{alggray}{// Update Interaction History}
  \ENDFOR
  \STATE \textcolor{algblue}{// \textbf{Answer Generation} (§\ref{sec:exploration_answer_generation})}
  \STATE $a \gets \textsc{GenerateAnswer}(q, \mathcal{N}, P_t, M_t)$
  \RETURN $a$
  \end{algorithmic}
\end{algorithm}

\section{Evaluation Dataset}
\label{sec:evaluation_dataset}

In this section, we provide detailed descriptions of the seven multi-hop question answering datasets used in our experiments, covering both KG-based and text-based multi-hop reasoning tasks.

\paragraph{KG-based QA Datasets.} These benchmarks primarily require reasoning over structured facts in the knowledge graph and usually expect entity-centric answers grounded in KG nodes or literal attributes.

\textbf{WebQSP}~\cite{WebQSP} is a widely used knowledge base QA benchmark derived from WebQuestions, containing questions paired with SPARQL queries over Freebase (mapped to Wikidata in our setting). It primarily focuses on questions that require 1-2 hop reasoning over structured KG facts to retrieve target entities.

\textbf{CWQ}~\cite{CWQ} extends the WebQSP by generating questions with higher complexity. It necessitates multi-hop reasoning (up to 4 hops) and challenges models to handle intricate logical constraints, such as composition, conjunctions, and superlatives over the graph structure.

\textbf{QALD10-en}~\cite{QALD10} is the English subset of the 10th Question Answering over Linked Data challenge, specifically adapted for Wikidata. It is characterized by highly complex natural language questions that map to intricate SPARQL queries. The dataset specifically tests a system's ability to handle advanced structural operations, such as aggregations, filters, and multi-hop relationships.

\textbf{KGQAGen}~\cite{KGQAGen} is a highly reliable, Wikidata-grounded benchmark generated via an LLM-in-the-loop framework with symbolic verification, featuring a wide range of topics and structural complexities. It systematically addresses the inaccurate annotations and ambiguity in existing evaluation datasets, making it a superior standard for evaluating multi-hop reasoning capabilities in KG-based RAG systems.

\paragraph{Text-based QA Datasets.} These benchmarks demand semantic understanding and multi-document synthesis, covering span extraction and more free-form answer types such as dates, numbers, boolean answers, and short descriptive phrases.

\textbf{2WikiMQA}~\cite{2WikiMQA} is a large-scale dataset synthesized from Wikidata and Wikipedia, specifically designed to test multi-hop reasoning across multiple documents. It focuses on questions that require comparison, aggregation, and bridging reasoning types. The questions typically require reasoning chains of 2 to 4 hops, challenging models to perform cross-document integration and logical inference over unstructured text.

\textbf{MusiQue}~\cite{MusiQue} is a highly challenging multi-hop QA dataset constructed by compositing single-hop questions in a bottom-up manner. It is explicitly designed to minimize reasoning shortcuts and enforce connected reasoning, ensuring that answering the final question strictly requires resolving intermediate sub-questions. The dataset features 2-4 hop questions that are significantly harder to solve via disconnected reasoning compared to HotpotQA or 2WikiMQA.

\textbf{AdvHotpotQA}~\cite{AdvHotpotQA} is an adversarial version of the  HotpotQA~\cite{HotpotQA} dataset. It introduces distractor paragraphs that share high lexical overlap with the question but do not contain the answer, rigorously testing the system's robustness against retrieval noise and its ability to identify supporting evidence accurately.

Table~\ref{tab:dataset_stats} summarizes the statistics of the evaluation datasets. For WebQSP, AdvHotpotQA, and QALD10-en, we use the same test splits as in ToG-2~\cite{ToG2} for fair comparison. For CWQ, we adopt the first 1,024 questions from the test set used in ToG~\cite{ToG}. For 2WikiMQA and MusiQue, we employ the evaluation sets provided by IRCoT~\cite{IRCoT}. For KGQAGen, we use the full test set from the KGQAGen-10k benchmark.

\begin{table}[h]
  \centering
  \renewcommand{\arraystretch}{1.1}
  \begin{tabular}{llc}
    \toprule
    \textbf{Category} & \textbf{Dataset} & \textbf{\# Questions} \\
    \midrule
    \multirow{4}{*}{\makecell{KG-based \\ QA}} 
    & WebQSP & 1,430 \\
    & CWQ & 1,024 \\
    & QALD10-en & 333 \\
    & KGQAGen-10k & 1,079 \\
    \midrule
    \multirow{3}{*}{\makecell{Text-based \\ QA}} 
    & 2WikiMQA & 500 \\
    & MusiQue & 500 \\
    & AdvHotpotQA & 308 \\
    \bottomrule
  \end{tabular}
  \caption{Statistics of the evaluation datasets.}
  \label{tab:dataset_stats}
\end{table}

\section{Knowledge Bases}
\label{sec:knowledge_bases}

CoG operates over a global knowledge environment that combines structured symbolic facts from Wikidata with entity-centric textual evidence from Wikipedia. This design allows the model to alternate between precise graph-based reasoning and flexible textual exploration, which is important for open-domain multi-hop questions where either source alone may be incomplete.

\paragraph{Structured Knowledge Graph (Wikidata).}
We use the full Wikidata dump\footnote{\url{https://dumps.wikimedia.org/wikidatawiki/entities/}} (May 2024) as the structured knowledge graph. After preprocessing, the resulting graph contains 89.3 million entities and 1.38 billion edges. We distinguish two types of edges: relational edges between entities (e.g., \textit{(Steve Jobs, founded, Apple)}), and attribute edges that connect entities to literal values such as dates, quantities, or strings (e.g., \textit{(Steve Jobs, date of birth, 1955-02-24)}). In total, the processed graph contains 823.7 million relational triples and 562.0 million attribute triples.

This large-scale setting is substantially different from task-specific or locally constructed KGs. It requires entity linking and retrieval to operate over a global entity space rather than a small candidate graph.

\paragraph{Knowledge Graph Preprocessing.}
We process Wikidata into a retrieval-oriented representation to efficiently support CoG's core operations.

For \textit{entity grounding}, we extract English labels, aliases, and descriptions. These are encoded via \texttt{Qwen3-Embedding-4B} into a FAISS index, enabling dense candidate recall across nearly 90 million entities with only $\sim$10GB of storage. Since many Wikidata entities share similar or identical names, each candidate is further supplemented with a compact entity profile, including textual metadata, popularity signals, and local graph context, for fine-grained LLM-based disambiguation.

For \textit{relation exploration and fact acquisition}, we organize relational and attribute triples into an in-memory key-value database (Redis) for efficient neighbor lookups. Additionally, we pre-compute the global occurrence frequencies of all relations, which serves as a crucial structural heuristic for the agent to prune high fan-out paths during exploration.

\paragraph{Unstructured Text Corpus (Wikipedia).} We employ the real-time English Wikipedia\footnote{\url{https://en.wikipedia.org/w/api.php}} as our unstructured text corpus. It comprises approximately 7.1 million entity-centric articles, totaling over 5 billion words. 
Distinct from traditional RAG that chunks text flatly, we process each page to retain its hierarchical structure, including the summary paragraph, table of contents, and sectioned full text. This structure supports CoG's hierarchical reading strategy (global skimming followed by detailed reading), enabling efficient navigation through long documents. 

Wikipedia complements Wikidata by providing richer natural-language context, especially for facts that are difficult to normalize into triples. In contrast, Wikidata provides normalized entities and typed relations for precise symbolic traversal. CoG exploits this complementarity by using KG facts as high-precision reasoning anchors and Wikipedia passages as high-recall contextual evidence.

\section{Additional Evaluation Metrics}
\label{sec:additional_metrics}

To complement the primary EM evaluation, Tables~\ref{tab:llm_judge_results} and~\ref{tab:f1_results} report LLM-as-judge \textsc{Correct} rates and token-level F1 scores, respectively, for the same model outputs as in Table~\ref{tab:main_results}.
For LLM-as-judge, Qwen3.5-122B-A10B (non-thinking mode, temperature 0) classifies each response as \textsc{Correct}, \textsc{Partial},
\textsc{Incorrect}, or \textsc{No Answer}, given only question, reference answers,
and model response. We conservatively report only the \textsc{Correct} rate. 
For token-level F1, the same model extracts a concise final answer using only the question and response. 
We then normalize predictions and references by lowercasing, removing punctuation and articles, and normalizing whitespace. 
For multiple references, we retain the maximum F1.

\begin{table*}[t]
  \centering
  \small
  \renewcommand{\arraystretch}{1.05}
  \resizebox{\textwidth}{!}{
  \begin{tabular}{llcccccccc}
      \toprule
      \multirow{2}{*}{\textbf{Baseline Type}} & \multirow{2}{*}{\textbf{Method}} & \multicolumn{4}{c}{\textbf{Multi-hop KG-based QA}} & \multicolumn{3}{c}{\textbf{Multi-hop text-based QA}} & \multirow{2}{*}{\textbf{Avg.}} \\
      \cmidrule(lr){3-6} \cmidrule(lr){7-9}
      & & KGQAGen & CWQ & QALD10-en & WebQSP & 2WikiMQA & AdvHotpotQA & MusiQue & \\
      \midrule
      \multirow{3}{*}{LLM-only}
      & Direct & 38.74 & 34.57 & 40.84 & 64.06 & 29.20 & 23.38 & 14.20 & 35.00 \\
      & CoT & 38.28 & 35.06 & 44.14 & 60.28 & 31.20 & 25.97 & 15.40 & 35.76 \\
      & CoT-SC & 39.48 & 32.71 & 43.54 & 61.05 & 32.40 & 26.62 & 14.80 & 35.80 \\
      \midrule
      \multirow{4}{*}{Text-based RAG}
      & Vector RAG & 50.23 & 33.20 & 37.84 & 64.83 & 48.80 & 45.45 & 18.20 & 42.65 \\
      & ReAct & 48.19 & 35.74 & 42.04 & 69.79 & 57.60 & 54.87 & 22.80 & 47.29 \\
      & IRCoT & 52.55 & 34.57 & 37.84 & 66.43 & 68.60 & 61.69 & 25.00 & 49.53 \\
      & Search-o1 & 52.27 & 39.84 & 48.65 & 68.60 & 47.80 & 47.73 & 27.40 & 47.47 \\
      \midrule
      \multirow{3}{*}{KG-based RAG}
      & ToG & 42.63 & 29.20 & 45.95 & 62.87 & 34.80 & 22.08 & 11.20 & 35.53 \\
      & PoG & 51.44 & 29.59 & 49.85 & 70.42 & 79.20 & 19.80 & 16.40 & 45.24 \\
      & DoG & 43.65 & 25.68 & 51.65 & 63.29 & 62.00 & 24.68 & 13.80 & 40.68 \\
      \midrule
      Hybrid RAG
      & ToG-$2$ & 53.57 & 41.50 & 49.85 & 70.21 & 65.20 & 42.86 & 22.40 & 49.37 \\
      \rowcolor{blue!10}
      \multirow{1}{*}{Proposed}
      & CoG & \textbf{66.82} & \textbf{44.92} & \textbf{59.76} & \textbf{76.57} & \textbf{88.20} & \textbf{62.34} & \textbf{32.80} & \textbf{61.63} \\
      \bottomrule
  \end{tabular}
  }
  \caption{LLM-as-judge \textsc{Correct} rate (\%) on seven datasets. The best result in each column is highlighted in \textbf{bold}.}
  \label{tab:llm_judge_results}
\end{table*}

\begin{table*}[t]
  \centering
  \small
  \renewcommand{\arraystretch}{1.05}
  \resizebox{\textwidth}{!}{
  \begin{tabular}{llcccccccc}
      \toprule
      \multirow{2}{*}{\textbf{Baseline Type}} & \multirow{2}{*}{\textbf{Method}} & \multicolumn{4}{c}{\textbf{Multi-hop KG-based QA}} & \multicolumn{3}{c}{\textbf{Multi-hop text-based QA}} & \multirow{2}{*}{\textbf{Avg.}} \\
      \cmidrule(lr){3-6} \cmidrule(lr){7-9}
      & & KGQAGen & CWQ & QALD10-en & WebQSP & 2WikiMQA & AdvHotpotQA & MusiQue & \\
      \midrule
      \multirow{3}{*}{LLM-only}
      & Direct & 38.64 & 33.17 & 28.14 & 49.59 & 33.19 & 21.63 & 15.77 & 31.45 \\
      & CoT & 40.28 & 35.25 & 34.35 & 55.96 & 35.94 & 25.13 & 17.68 & 34.94 \\
      & CoT-SC & 41.91 & 34.59 & 35.29 & 56.30 & 36.14 & 24.58 & 18.11 & 35.27 \\
      \midrule
      \multirow{4}{*}{Text-based RAG}
      & Vector RAG & 48.49 & 28.98 & 24.43 & 47.81 & 44.61 & 36.84 & 17.50 & 35.52 \\
      & ReAct & 46.15 & 29.90 & 25.58 & 49.78 & 51.55 & 42.78 & 21.99 & 38.25 \\
      & IRCoT & 51.26 & 30.34 & 25.19 & 48.08 & 59.86 & 47.93 & 23.13 & 40.83 \\
      & Search-o1 & 49.18 & 35.43 & 31.39 & 51.48 & 44.22 & 39.44 & 25.65 & 39.54 \\
      \midrule
      \multirow{3}{*}{KG-based RAG}
      & ToG & 43.31 & 27.24 & 35.58 & 58.21 & 34.31 & 19.51 & 13.22 & 33.05 \\
      & PoG & 51.30 & 26.46 & 39.84 & 59.35 & 73.47 & 16.47 & 15.56 & 40.35 \\
      & DoG & 46.00 & 27.20 & \textbf{43.20} & 56.01 & 61.84 & 21.15 & 14.35 & 38.54 \\
      \midrule
      Hybrid RAG
      & ToG-$2$ & 54.96 & \textbf{39.46} & 38.55 & \textbf{60.58} & 62.80 & 37.85 & 23.45 & 45.38 \\
      \rowcolor{blue!10}
      \multirow{1}{*}{Proposed}
      & CoG & \textbf{67.69} & 38.46 & 37.96 & 50.36 & \textbf{76.22} & \textbf{49.95} & \textbf{28.48} & \textbf{49.87} \\
      \bottomrule
  \end{tabular}
  }
  \caption{Token-level F1 (\%) on seven datasets. The best result in each column is highlighted in \textbf{bold}.}
  \label{tab:f1_results}
\end{table*}

As shown in Tables~\ref{tab:llm_judge_results} and~\ref{tab:f1_results}, the complementary metrics are consistent with the EM-based findings. Under LLM-as-judge, CoG achieves the best result on all seven datasets, with an average \textsc{Correct} rate of 61.63\%, outperforming the strongest baseline, IRCoT, by 12.10 points. Under token-level F1, CoG also obtains the highest average score of 49.87\%, improving over ToG-2 by 4.49 points. These results show that CoG's gains are not merely caused by exact string matching, but remain clear under both semantic equivalence judgment and lexical-overlap evaluation.

\section{Detailed Results on Backbone LLMs}
\label{sec:appendix_llm_analysis}
This section supplements the backbone LLM generalizability analysis in Section~\ref{sec:in-depth_analysis}. Table~\ref{tab:full_backbone_analysis} presents the detailed performance breakdown across different backbone LLMs, corresponding to the summarized results in Table~\ref{tab:llm_generalization}.

\paragraph{Evaluation Setup.} 
To balance computational cost with evaluation rigor, we adopted the following dataset selection and sampling strategies:

\textbf{Dataset Selection.} We excluded the WebQSP dataset from this analysis, as modern LLMs already achieve near-saturated performance on its simple 1-2 hop queries in the Direct Prompting setting, making it less discriminative for measuring improvement.
    
\textbf{Open-source Models.} For large-scale KG-based datasets (KGQAGen and CWQ), we sampled 300 instances each to align with the scale of QALD10-en. For the remaining datasets (QALD10-en, 2WikiMQA, AdvHotpotQA, and MusiQue), we evaluated on the full test sets.
    
\textbf{Proprietary Models.} We evaluated on a sampled subset of 100 examples for each dataset.

\paragraph{Key Observations.}
The detailed results reveal three key insights regarding CoG's generalizability and cost-effectiveness:
(1) \textbf{Broad Generalizability:} CoG consistently improves performance across all backbone models, regardless of their scale (from 8B to 235B) or type (open-source vs. proprietary).
(2) \textbf{Unlocking Small Models:} The relative performance gain is most pronounced on the smallest model, Qwen3-8B (+113.2\%), showing that CoG's cognitive architecture effectively compensates for the limited parametric knowledge of smaller LLMs.
(3) \textbf{Cost-Effectiveness:} CoG-8B (Avg. 51.0\%) outperforms direct prompting on massive models (e.g. Qwen3-235B 43.7\%, DeepSeek-V3.2 38.0\%) and surpasses strong 32B baselines (e.g., IRCoT 44.6\%, ToG-2 45.9\%). Since 8B models are cheaper and faster, CoG-8B offers a highly cost-effective deployment solution that neutralizes the token overhead of multi-turn exploration.

\begin{table*}[t]
  \centering
  \small
  \renewcommand{\arraystretch}{1.1}
  \resizebox{\textwidth}{!}{
  \begin{tabular}{llcccccc|ccc}
      \toprule
      \multirow{2}{*}{\textbf{Backbone}} & \multirow{2}{*}{\textbf{Method}} & \multicolumn{3}{c}{\textbf{KG-based QA}} & \multicolumn{3}{c|}{\textbf{Text-based QA}} & \multicolumn{3}{c}{\textbf{Average}} \\
      \cmidrule(lr){3-5} \cmidrule(lr){6-8} \cmidrule(lr){9-11}
      & & KGQAGen & CWQ & QALD & 2Wiki & Hotpot & MusiQ & \textbf{KG} & \textbf{Text} & \textbf{Total} \\
      \midrule
      \rowcolor{gray!12}
      \multicolumn{11}{c}{\textit{Open-Source Models}} \\
      \multirow{6}{*}{Qwen3-32B} 
      & Direct & 40.00 & 37.00 & 47.45 & 47.20 & 20.13 & 8.40 & 41.48 & 25.24 & 33.36 \\
      & IRCoT & 53.80 & 33.30 & 41.10 & 71.60 & 47.10 & 20.60 & 42.73 & 46.43 & 44.58 \\
      & Search-o1 & 46.34 & 36.72 & 43.54 & 44.60 & 40.26 & 19.40 & 42.20 & 34.75 & 38.48 \\
      & DoG & 54.80 & 33.10 & \textbf{55.90} & 61.20 & 20.10 & 11.40 & 47.93 & 30.90 & 39.42 \\
      & ToG-2 & 55.60 & 41.40 & 53.50 & 73.00 & 35.70 & 16.20 & 50.17 & 41.63 & 45.90 \\
      & CoG & \textbf{76.33} & \textbf{46.33} & 55.26 & \textbf{85.20} & \textbf{51.95} & \textbf{27.80} & \textbf{59.31} & \textbf{54.98} & \textbf{57.15} \\
      \midrule
      \multirow{2}{*}{Qwen3-8B} 
      & Direct & 29.10 & 27.83 & 36.34 & 31.80 & 12.99 & 5.40 & 31.09 & 16.73 & 23.91 \\
      & CoG & \textbf{68.03} & \textbf{42.29} & \textbf{49.85} & \textbf{80.00} & \textbf{45.45} & \textbf{20.20} & \textbf{53.39} & \textbf{48.55} & \textbf{50.97} \\
      \midrule
      \multirow{2}{*}{Qwen3-235B} 
      & Direct & 54.67 & 44.33 & 54.95 & 55.20 & 32.47 & 20.80 & 51.32 & 36.16 & 43.74 \\
      & CoG & \textbf{79.00} & \textbf{55.11} & \textbf{56.46} & \textbf{87.60} & \textbf{54.55} & \textbf{32.80} & \textbf{63.52} & \textbf{58.32} & \textbf{60.92} \\
      \midrule
      \multirow{2}{*}{gpt-oss-120b} 
      & Direct & 18.67 & 29.67 & 36.64 & 20.80 & 18.83 & 8.00 & 28.33 & 15.88 & 22.10 \\
      & CoG & \textbf{42.33} & \textbf{40.33} & \textbf{45.95} & \textbf{63.40} & \textbf{38.64} & \textbf{18.80} & \textbf{42.87} & \textbf{40.28} & \textbf{41.58} \\
      \midrule
      \multirow{2}{*}{DeepSeek-V3.2} 
      & Direct & 53.67 & 45.33 & 44.74 & 39.20 & 31.17 & 13.80 & 47.91 & 28.06 & 37.99 \\
      & CoG & \textbf{85.33} & \textbf{51.67} & \textbf{57.36} & \textbf{88.40} & \textbf{59.09} & \textbf{34.00} & \textbf{64.79} & \textbf{60.50} & \textbf{62.64} \\
      \midrule
      \rowcolor{gray!12}
      \multicolumn{11}{c}{\textit{Proprietary Models (Sampled Set, N=100)}} \\
      \multirow{2}{*}{GPT-5.1} 
      & Direct & 54.00 & 47.00 & 36.00 & 59.00 & 40.00 & 17.00 & 45.67 & 38.67 & 42.17 \\
      & CoG & \textbf{81.00} & \textbf{49.00} & \textbf{38.00} & \textbf{82.00} & \textbf{55.00} & \textbf{29.00} & \textbf{56.00} & \textbf{55.33} & \textbf{55.67} \\
      \midrule
      \multirow{2}{*}{Gemini-2.5-Flash} 
      & Direct & 62.00 & \textbf{46.00} & 27.00 & 65.00 & 46.00 & 22.00 & 45.00 & 44.33 & 44.67 \\
      & CoG & \textbf{73.00} & 44.00 & \textbf{31.00} & \textbf{78.00} & \textbf{58.00} & \textbf{23.00} & \textbf{49.33} & \textbf{53.00} & \textbf{51.17} \\
      \bottomrule
  \end{tabular}
  }
  \caption{Detailed performance comparison (EM, \%) of different backbone LLMs with and without the CoG framework across six datasets (WebQSP excluded). For Qwen3-32B, we additionally include strong baselines from various categories (Text-based, KG-based, and Hybrid RAG) to demonstrate that CoG with a lightweight 8B model can outperform these 32B baselines.}
  \label{tab:full_backbone_analysis}
\end{table*}

\section{Detailed Results on Exploration Depth}
\label{sec:appendix_depth_analysis}

Table~\ref{tab:depth_detailed} provides the full results corresponding to the exploration depth analysis discussed in Section~\ref{sec:in-depth_analysis} and visualized in Figure~\ref{fig:depth_analysis}. The table lists the Exact Match (EM) scores for CoG across exploration depths from 1 to 6 on all seven benchmark datasets, alongside the average performance on KG-based and Text-based tasks.

\begin{table*}[h]
  \centering
  \small
  \renewcommand{\arraystretch}{1.2}
  \resizebox{\textwidth}{!}{
  \begin{tabular}{lccccc|cccc|c}
      \toprule
      \multirow{2}{*}{\textbf{Method}} & \multicolumn{5}{c|}{\textbf{KG-based QA}} & \multicolumn{4}{c|}{\textbf{Text-based QA}} & \multirow{2}{*}{\textbf{Avg.}} \\
      & KGQAGen & CWQ & QALD10 & WebQSP & \textbf{Avg.} & 2Wiki & AdvHotpot & MusiQue & \textbf{Avg.} & \\
      \midrule
      VectorRAG & 48.66 & 30.76 & 39.04 & 70.49 & 47.24 & 52.60 & 32.14 & 12.40 & 32.38 & 40.87 \\
      \midrule
      CoG (Depth=1) & 67.47 & 37.99 & 45.35 & 80.63 & 57.86 & 48.00 & 26.62 & 4.60 & 26.41 & 44.38 \\
      CoG (Depth=2) & 71.83 & 42.58 & 51.95 & 81.05 & 61.85 & 85.20 & 46.43 & 20.00 & 50.54 & 57.01 \\
      CoG (Depth=3) & 74.24 & 42.58 & 53.75 & \textbf{81.75} & 63.08 & \textbf{86.80} & 49.35 & 25.00 & 53.72 & 59.07 \\
      CoG (Depth=4) & 74.70 & 42.87 & \textbf{55.26} & \textbf{81.75} & \textbf{63.65} & 85.20 & 51.95 & \textbf{27.80} & \textbf{54.98} & \textbf{59.93} \\
      CoG (Depth=5) & 74.51 & 42.97 & 51.65 & 81.47 & 62.65 & 84.00 & 52.92 & 26.00 & 54.31 & 59.07 \\
      CoG (Depth=6) & \textbf{75.90} & \textbf{43.65} & 54.35 & 80.56 & 63.62 & 83.20 & \textbf{53.25} & 25.60 & 54.02 & 59.50 \\
      \bottomrule
  \end{tabular}
  }
  \caption{Full results of exploration depth analysis. We report Exact Match (EM) scores for CoG with maximum exploration depths ranging from 1 to 6 on all seven datasets, alongside the VectorRAG baseline. The best result for each dataset is highlighted in \textbf{bold}.}
  \label{tab:depth_detailed}
\end{table*}

\section{Detailed Efficiency Profile}
\label{sec:appendix_efficiency}

This section details CoG's computational overhead across token consumption, end-to-end latency, and wall-clock time per retrieval round.

\begin{table*}[t]
  \centering
  \small
  \renewcommand{\arraystretch}{1.1}
  \resizebox{\textwidth}{!}{
  \begin{tabular}{l|cc|cc|cc|cc|cc|cc|cc}
      \toprule
      \multirow{2}{*}{\textbf{Method}} & \multicolumn{2}{c|}{\textbf{KGQAGen}} & \multicolumn{2}{c|}{\textbf{CWQ}} & \multicolumn{2}{c|}{\textbf{QALD10-en}} & \multicolumn{2}{c|}{\textbf{WebQSP}} & \multicolumn{2}{c|}{\textbf{2WikiMQA}} & \multicolumn{2}{c|}{\textbf{AdvHotpotQA}} & \multicolumn{2}{c}{\textbf{MusiQue}} \\
      \cmidrule(lr){2-3} \cmidrule(lr){4-5} \cmidrule(lr){6-7} \cmidrule(lr){8-9} \cmidrule(lr){10-11} \cmidrule(lr){12-13} \cmidrule(lr){14-15}
      & Token & Latency & Token & Latency & Token & Latency & Token & Latency & Token & Latency & Token & Latency & Token & Latency \\
      \midrule
      ReAct & 3,850 & 14.1 & 3,996 & 13.7 & 3,908 & 13.8 & 3,870 & 12.3 & 3,984 & 18.8 & 3,631 & 13.2 & 7,758 & 20.0 \\
      IRCoT & 23,832 & 32.9 & 26,107 & 28.9 & 23,483 & 33.7 & 22,798 & 19.5 & 22,350 & 41.5 & 23,038 & 38.2 & 29,798 & 46.1 \\
      Search-o1 & 12,262 & 98.4 & 13,505 & 119.8 & 10,611 & 84.0 & 9,891 & 62.6 & 19,196 & 140.5 & 11,932 & 77.9 & 20,913 & 120.0 \\
      ToG & 6,552 & 68.7 & 4,168 & 41.9 & 4,154 & 69.7 & 2,963 & 39.1 & 4,779 & 70.5 & 4,946 & 68.4 & 4,695 & 75.2 \\
      PoG & 8,752 & 16.7 & 5,117 & 12.7 & 3,660 & 10.9 & 2,862 & 7.0 & 4,909 & 16.6 & 7,220 & 23.7 & 11,899 & 28.4 \\
      DoG & 15,494 & 168.7 & 10,304 & 133.9 & 3,088 & 20.6 & 3,320 & 18.1 & 6,671 & 105.4 & 10,364 & 158.7 & 8,575 & 134.0 \\
      ToG-2 & 4,193 & 109.7 & 3,424 & 78.6 & 3,273 & 53.3 & 3,127 & 51.3 & 3,067 & 42.7 & 3,398 & 51.0 & 4,008 & 95.0 \\
      \midrule
      \rowcolor{blue!10}
      \textbf{CoG} & 36,323 & 285.1 & 29,465 & 194.2 & 15,636 & 95.7 & 14,204 & 80.7 & 15,508 & 121.8 & 18,958 & 145.8 & 38,876 & 263.5 \\
      \bottomrule
  \end{tabular}
  }
  \caption{Token consumption and latency (seconds) per question across different methods. Values represent the median of the dataset to reflect the typical cost for a standard query.}
  \label{tab:token_latency}
\end{table*}

\textbf{Complexity Adaptability.} CoG dynamically scales resources to query difficulty (Table~\ref{tab:token_latency}). For simpler tasks (WebQSP, 2WikiMQA), it uses fewer tokens ($\sim$14-15k) than IRCoT ($\sim$22k) yet achieves higher accuracy. For complex tasks (MusiQue, KGQAGen), it invests more tokens into deep reasoning, yielding substantial gains where efficient methods fail (+11.6\% on MusiQue over ToG-2).

\begin{figure}[h]
  \centering
  \includegraphics[width=\columnwidth]{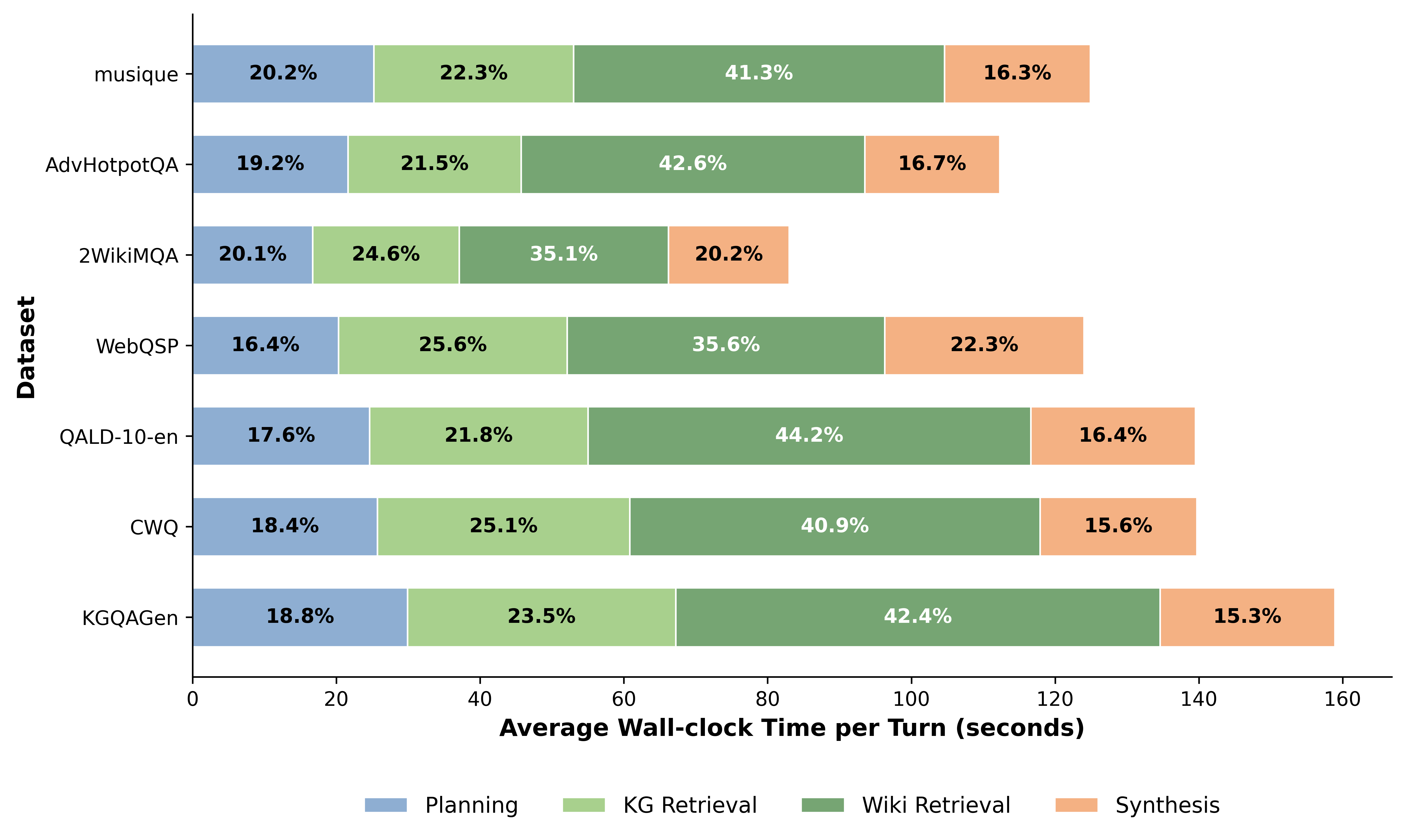}
  \caption{Average wall-clock time (seconds) breakdown per retrieval round across different datasets.}
  \label{fig:time_breakdown}
\end{figure}

\textbf{Latency Breakdown.} Figure~\ref{fig:time_breakdown} details the average wall-clock time per retrieval round. The latency primarily stems from sequential LLM calls in the cognitive cycle. However, this extended reasoning time supports a fully open-ended retrieval setting: CoG starts from the raw question without pre-annotated entities, directly grounds and retrieves evidence, and adaptively adjusts KG/text exploration according to the evolving reasoning state. Furthermore, this deep filtering ensures precise localization, reducing accessed documents by $12.4\times$ compared to ToG-2 (Table~\ref{tab:doc_efficiency}). The current wall-clock time can be further optimized by parallelizing the dual-source exploration modules and concurrent sub-queries.

\textbf{Cost-Effectiveness.} While CoG consumes more tokens on complex tasks, this is offset by its ability to utilize smaller, cheaper models. As shown in Appendix~\ref{sec:appendix_llm_analysis}, CoG with an 8B model outperforms strong 32B baselines, suggesting a more favorable accuracy-cost trade-off for practical deployment.

\section{Sensitivity Analysis on Entity Linking}
\label{sec:appendix_el_sensitivity}

To evaluate the robustness of CoG to the candidate retrieval size (top-$k$) in the entity linking module, we vary the number of candidate entities from $6$ to $30$.
We conduct this analysis on four representative datasets, covering both KG-based and text-based QA tasks.
For KGQAGen, we evaluate on a 300-sample subset to control computational cost.

\begin{table}[!t]
  \centering
  \small
  \resizebox{\columnwidth}{!}{
  \begin{tabular}{lccccc}
      \toprule
      \textbf{$k$} & \textbf{KGQAGen} & \textbf{2Wiki} & \textbf{Hotpot} & \textbf{MusiQ} & \textbf{Avg.} \\
      \midrule
      6  & 79.3 & 84.8 & 47.7 & 25.6 & 59.4 \\
      12 & 78.3 & 84.2 & 49.7 & 25.2 & 59.4 \\
      18 & 74.3 & \textbf{85.8} & 50.0 & 26.2 & 59.1 \\
      20 & 76.3 & 85.2 & \textbf{52.0} & \textbf{27.8} & \textbf{60.3} \\
      30 & 78.0 & 85.6 & 51.3 & 26.2 & \textbf{60.3} \\
      \bottomrule
  \end{tabular}
  }
  \caption{Sensitivity analysis of entity-linking candidate size $k$. We report EM (\%) on four representative datasets.}
  \label{tab:el_sensitivity}
\end{table}

As shown in Table~\ref{tab:el_sensitivity}, CoG is robust to the entity-linking candidate size.
Across $k=6$ to $30$, the average EM remains within a narrow range of 59.1\%--60.3\%.
Even with a narrow retrieval scope ($k=6$), CoG achieves a strong average EM of 59.4\%, while increasing $k$ provides broader entity recall without causing overall performance degradation.
We set $k=20$ as the default because it achieves the best average performance, tied with $k=30$, while using fewer candidate entities.
This indicates that CoG's context-aware disambiguation effectively filters noisy candidates while benefiting from broader entity recall.

\section{Manual Evaluation and Error Study}
\label{sec:manual_evaluation}

As the Exact Match (EM) metric strictly relies on string matching, it often fails to evaluate the validity of the underlying reasoning process. To gain deeper insights, we conduct a manual analysis on a sampled subset. We assess \textbf{Reasoning Faithfulness} for cases where the EM score is correct (EM=1), and perform an \textbf{Error Diagnosis} for cases where the model is judged incorrect (EM=0).

\subsection{Reasoning Faithfulness}
\label{sec:reasoning_faithfulness}

For predictions judged correct by EM, we further examine whether the model reaches the answer through a faithful reasoning process. We categorize each EM-correct prediction into four types:
\begin{itemize}[leftmargin=1.5em, topsep=4pt, itemsep=2pt, parsep=0pt]
    \item \textbf{True Positive (TP):} The reasoning chain is logically sound, and the extracted evidence fully supports the final answer.
    \item \textbf{Flawed True Positive (Flawed TP):} The final answer and core reasoning direction are correct, but the reasoning contains minor verification gaps, shortcut steps, or non-critical factual imperfections.
    \item \textbf{Weak False Positive (Weak FP):} The final answer is correct by coincidence, but the core intermediate entities or facts are substantially incorrect, i.e., the model is right for the wrong reasons.
    \item \textbf{False Positive (FP):} The model fails to deduce the correct answer, but the EM metric falsely judges it correct due to substring matching in the reasoning text.
\end{itemize}

Table~\ref{tab:reasoning_faithfulness} shows that CoG produces the most faithful EM-correct predictions. Its severe false-positive rate (Weak FP + FP) is only 3.6\%, much lower than ToG-2 (12.7\%) and Self-Consistency (29.1\%). Even when Flawed TP is treated as non-ideal reasoning, CoG remains the most reliable method.

Self-Consistency is vulnerable to ungrounded parametric reasoning, often producing hallucinated bridge entities or fabricated attributes (\textit{Weak FPs}), or long CoT traces that accidentally match the gold string (\textit{FPs}). ToG-2 reduces such hallucinations through retrieval, but its reactive exploration can still preserve wrong intermediate entities or unresolved constraints. CoG's graph-text exploration and reflection better ground intermediate steps, so most of its imperfect cases involve incomplete verification of secondary constraints (\textit{Flawed TPs}) rather than broken core reasoning chains.

\begin{table*}[!t]
  \centering
  \small
  \begin{tabular}{lcccc}
      \toprule
      \textbf{Method} & \textbf{TP} & \textbf{Flawed TP} & \textbf{Weak FP} & \textbf{FP} \\
      \midrule
      Self-Consistency & 60.0\% & 10.9\% & 21.8\% & 7.3\% \\
      ToG-2 & 81.8\% & 5.5\% & 3.6\% & 9.1\% \\
      \textbf{CoG (Ours)} & 89.1\% & 7.3\% & 1.8\% & 1.8\% \\
      \bottomrule
  \end{tabular}
  \caption{Distribution of reasoning faithfulness for predictions evaluated as correct by EM.}
  \label{tab:reasoning_faithfulness}
\end{table*}

Figure~\ref{fig:case_box} further shows that identical EM scores can hide different reasoning quality. While SC and ToG-2 may reach the correct answer through hallucinated or weakly supported intermediate links, CoG derives the answer from a more grounded and faithful reasoning path. This example also involves a temporal update: although the dataset reference path is based on an older coach, CoG identifies the updated coach and reaches the  correct nationality through verifiable evidence.

\begin{figure}[!t]
\centering
\fbox{%
  \begin{minipage}{\dimexpr\columnwidth-2\fboxsep-2\fboxrule\relax}
  \small
  \textbf{Question:} \textit{What nationality is the sport club that the current coach of Werder Bremen played professionally for?} \\
  \textbf{Gold Answer:} German \\
  \textbf{Reference Reasoning Path:} \\
  \textit{Werder Bremen $\rightarrow$ (coach) Alexander Nouri $\rightarrow$ (played for) VfL Osnabrück $\rightarrow$ (nationality) German}

  \vspace{0.5em}
  \textbf{\textcolor{red}{\ding{55}} SC (Weak FP):} \\
  \textit{Werder Bremen $\rightarrow$ Florian Kohfeldt $\rightarrow$ \textbf{\textcolor{red}{[Hallucinated]}} VfL Wolfsburg $\rightarrow$ German}\\
  Hallucinates that coach F. Kohfeldt played for VfL Wolfsburg (he only coached them). Hits the answer ``German'' purely by coincidence.

  \vspace{0.5em}
  \textbf{\textcolor{red}{\ding{55}} ToG-2 (Weak FP):} \\
  \textit{Werder Bremen $\rightarrow$ Florian Kohfeldt $\rightarrow$ \textbf{\textcolor{red}{[Guessed]}} Werder Bremen $\rightarrow$ German}\\
  Fails to retrieve his actual former clubs. Circularly assumes that since he coaches Werder Bremen (a German club), he must have played for them, guessing ``German''.

  \vspace{0.5em}
  \textbf{\textcolor{teal}{\ding{51}} CoG (True Positive):} \\
  \textit{Werder Bremen $\rightarrow$ Daniel Thioune $\rightarrow$ VfL Osnabrück $\rightarrow$ German}\\
  Identifies current coach D. Thioune from updated knowledge. Retrieves his actual former clubs (e.g., VfL Osnabrück) and confirms they are German.
  \end{minipage}%
}
\caption{A qualitative example showing CoG's superior reasoning faithfulness despite identical EM scores across models.}
\label{fig:case_box}
\end{figure}

\subsection{Error Diagnosis}
\label{sec:error_diagnosis}
\begin{figure}[h]
  \centering
  \includegraphics[width=\linewidth]{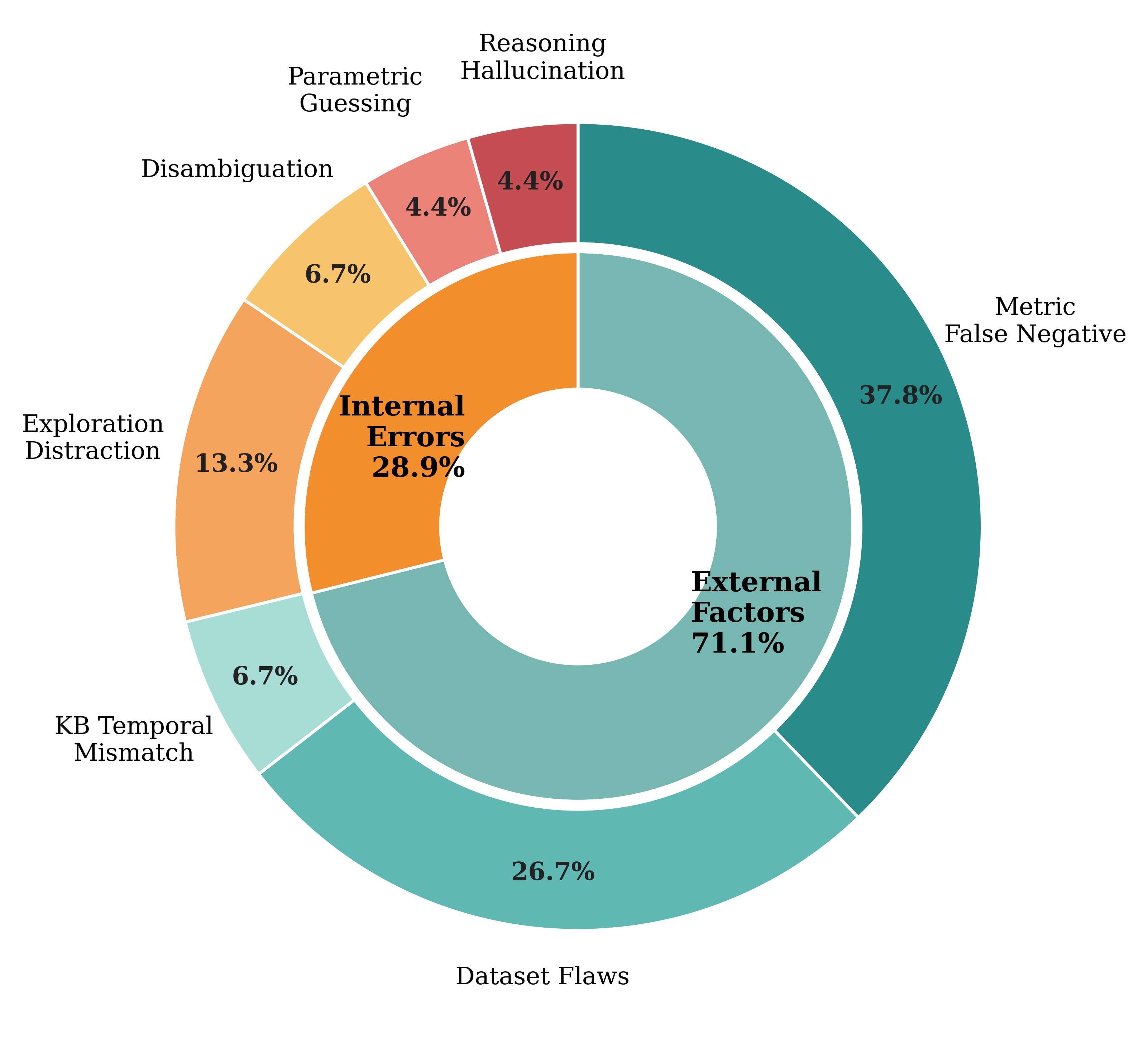}
  \caption{Taxonomy of error causes for CoG based on manual analysis.}
  \label{fig:manual_error_analysis}
\end{figure}

To understand the limitations of CoG, we conduct a primary-cause error diagnosis on a sampled subset of instances where CoG fails the EM evaluation. As illustrated in Figure~\ref{fig:manual_error_analysis}, we categorize the errors into External Factors and Internal Errors. 

\paragraph{External Factors (71.1\%):} A large portion of the errors are attributable to benchmark or evaluation limitations.
\begin{itemize}[leftmargin=1.5em, topsep=4pt, itemsep=2pt, parsep=0pt]
    \item \textbf{Metric False Negative (37.8\%):} The model successfully completes the multi-hop reasoning but fails the strict string matching. Common instances include  surface-form mismatch (aliases or pseudonyms), paraphrases, or answer granularity differences (city vs. metro area).
    \item \textbf{Dataset Flaws (26.7\%):} These include typos in the questions, grammatically ambiguous prompts leading to misaligned relation extraction, and noisy ground-truth annotations.
    \item \textbf{KB Temporal Mismatch (6.7\%):} Temporal mismatch occurs when the model retrieves updated facts from the knowledge source (e.g., recent sports team transfers), which conflict with the outdated static dataset ground truth.
\end{itemize}

\paragraph{Internal Errors (28.9\%):} The remaining errors represent genuine limitations in the agentic exploration process.
\begin{itemize}[leftmargin=1.5em, topsep=4pt, itemsep=2pt, parsep=0pt]
    \item \textbf{Exploration Distraction (13.3\%):} In environments with long texts or ambiguous clues, the agent occasionally gets trapped by irrelevant entities or over-thinks historical details, exhausting its maximum exploration turns.
    \item \textbf{Disambiguation (6.7\%):} The model sometimes overlooks rigid constraints during entity disambiguation (e.g., ignoring a "state school" constraint and erroneously selecting a private school).
    \item \textbf{Reasoning Hallucination (4.4\%) \& Parametric Guess (4.4\%):} The model occasionally struggles with counting logic over complex long tables, or bypasses retrieval entirely by relying on over-confident parametric guesses.
\end{itemize}

Overall, this diagnosis shows that raw EM can underestimate CoG, with many failures caused by rigid string matching, noisy annotations, or temporal mismatch. The remaining internal errors highlight important directions for improvement in exploration focus, entity disambiguation, and evidence-grounded verification.

\section{Case Studies}
\label{sec:case_studies}

Table~\ref{tab:case_tog} and Table~\ref{tab:case_cog} compare the reasoning trajectories of ToG-2 and CoG on a complex multi-hop question from AdvHotpotQA.

\paragraph{ToG-2 Analysis.} It fails due to its passive, graph-driven exploration strategy. The system starts by retrieving general information about ``Stanford Cardinal'' but fails to locate the specific player due to the absence of statistical attributes in the KG. Lacking a reflection mechanism, ToG-2 cannot diagnose this failure. Instead, it blindly continues its beam search, pivoting to high-ranking but irrelevant entities like ``quarterback'' in Turn 3, ultimately leading to an irreversible dead end. 

\paragraph{CoG Analysis.} In contrast, CoG demonstrates robust navigation through its cognitive cycle and bidirectional synergy. CoG initially encounters the same dead end as ToG-2 in Turn 1. However, Reflection diagnoses the strategy as too narrow. Proactive Planning then pivots to broader queries like ``football receiving leaders'' instead of the player directly (Turn 2). Crucially, Text Entity Utilization bridges KG gaps by identifying the entity ``James Lofton'', which aligns with the statistical constraints in the question, from Wikipedia related sections. This bidirectional synergy enables CoG to dynamically identify the target player and verify the specific statistics in Turn 3, successfully deriving the answer ``David''.

\begin{table*}[t]
  \centering
  \small
  \begin{tabularx}{\textwidth}{p{1.5cm}|X}
  \toprule
  \multicolumn{2}{p{\dimexpr\textwidth-2\tabcolsep\relax}}{\textbf{Question:} \textit{What is the middle name of the former Stanford Cardinals player who caught 57 passes for 1010 yards and 14 TDs?}} \\
  \multicolumn{2}{p{\dimexpr\textwidth-2\tabcolsep\relax}}{\textbf{Pre-annotated Entities:} Stanford Cardinal (Q2938060), Touchdown (Q650807)} \\
  \multicolumn{2}{p{\dimexpr\textwidth-2\tabcolsep\relax}}{\textbf{Ground Truth:} David} \\
  \midrule
  \textbf{Turn} & \textbf{ToG-2 Reasoning Process} \\
  \midrule
  \textbf{turn 1} & 
  \textbf{Starting Entities:} Stanford Cardinal (Q2938060), Touchdown (Q650807) \\
  & \textbf{Relations Explored:} member of sports team, sport, educated at, facet of, named after (6 relations) \\
  & \textbf{Retrieved Entities:} 57 entities (49 from ``member of sports team'', 4 from ``sport'', 1 from ``educated at'', 2 from ``Touchdown'' relations) \\
  & \quad Notable: 49 Stanford athletes including Greg Camarillo, John Elway, Mark Marquess, etc. \\
  & \textbf{Document Retrieval:} Retrieved Wikipedia pages for 55 entities (2 pages failed: sports terminology, Margaret Hayden) \\
  & \cellcolor{red!15}\textbf{Issue:} No specific statistics matching 57/1010/14 found in documents \\
  & \textbf{Reasoning:} Information insufficient - cannot identify the specific player \\
  & \textbf{Entity Ranking (for next turn):} \\
  & \quad [0.28] Greg Camarillo | [0.12] John Elway | [0.05] Mark Marquess \\
  \midrule
  \textbf{turn 2} & 
  \textbf{Starting Entities:} Greg Camarillo (Q1248387), John Elway (Q432931), Mark Marquess (Q6768732) \\
  & \textbf{Relations Explored:} given name, educated at, member of sports team, position played, occupation (9 relations) \\
  & \textbf{Retrieved Entities:} 11 entities (Greg, Menlo-Atherton HS, Stanford Univ., 5 NFL teams, John, quarterback, Mark, baseball player, Stanford Cardinal) \\
  & \textbf{Document Retrieval:} Retrieved Wikipedia pages for all 11 entities \\
  & \cellcolor{red!15}\textbf{Issue:} Found Greg Camarillo's given name but no middle name; no confirmation of statistics \\
  & \textbf{Reasoning:} Cannot confirm Greg Camarillo is the target player or find his middle name \\
  & \textbf{Entity Ranking (for next turn):} \\
  & \quad [0.45] Stanford Cardinal | [0.00] quarterback \\
  \midrule
  \textbf{turn 3} & 
  \textbf{Starting Entities:} Stanford Cardinal (Q2938060), quarterback (Q622747) \\
  & \textbf{Relations Explored:} member of sports team, position played, sport (6 relations) \\
  & \textbf{Retrieved Entities:} 106 entities (50 from ``member of sports team'', 4 from ``sport'', 1 from ``educated at'', 50 from ``position played'', 1 from ``part of'') \\
  & \quad Notable: 50 quarterbacks (Logan Thomas, John Brantley, Charlie Whitehurst, etc.), 50 Stanford athletes \\
  & \textbf{Document Retrieval:} Retrieved Wikipedia pages for 103 entities (3 pages failed: Cameron Dantley, Dai Minagawa, Margaret Hayden) \\
  & \cellcolor{red!15}\textbf{Critical Failure:} Explored wrong position (quarterback vs. wide receiver) \\
  & \cellcolor{red!15}\textbf{No Reflection:} System continued beam search without recognizing strategic misalignment \\
  & \textbf{Entity Ranking:} [0.31] Logan Thomas | [0.12] John Brantley | [0.02] Charlie Whitehurst \\
  & \textbf{Final Reasoning:} Information insufficient to identify the player or middle name \\
  & \cellcolor{red!15}\textbf{Outcome:} \textbf{FAILURE} - Generated answer without evidence \\
  \bottomrule
  \end{tabularx}
  \caption{Reasoning trace of ToG-2. The method employs a rigid beam search where the next exploration steps are determined by ranking score based on entity document relevance. Without reflection, the pruning mechanism blindly discards potentially correct paths and misguides the search towards irrelevant high-ranking entities, leading to an irreversible failure. Red highlights indicate critical failure analysis for the system.}
  \label{tab:case_tog}
\end{table*}
  
\begin{table*}[!t]
  \centering
  \small
  \begin{tabularx}{\textwidth}{p{1.5cm}|X}
  \toprule
  \multicolumn{2}{p{\dimexpr\textwidth-2\tabcolsep\relax}}{\textbf{Question:} \textit{What is the middle name of the former Stanford Cardinals player who caught 57 passes for 1010 yards and 14 TDs?}} \\
  \multicolumn{2}{p{\dimexpr\textwidth-2\tabcolsep\relax}}{\textbf{Ground Truth:} David} \\
  \midrule
  \textbf{Turn} & \textbf{CoG Reasoning Process} \\
  \midrule
  \textbf{Turn 1} & 
  \textbf{Planning Phase:} \\
  & \quad Analysis: ``To determine the middle name, I first need to identify the specific player who caught 57 passes for 1010 yards and 14 touchdowns.'' \\
  & \quad Query: ``Stanford Cardinals player with 57 catches, 1010 yards, 14 TDs'' \\
  & \quad Entity: ``Stanford Cardinals player'' \\
  & \textbf{Exploration Phase:} \\
  & \quad \textit{KG:} Entity linking failed - 9 candidates all team-level entities (e.g., Stanford Cardinal football, Stanford Cardinal men's soccer), no individual player entities \\
  & \quad \textit{Wikipedia:} Searched ``Stanford Cardinal football'' page, extracted team context (Jim Plunkett, Andrew Luck) but no player matching the statistics \\
  & \cellcolor{cyan!20}\textbf{Synthesis:} No player-specific information found \\
  & \cellcolor{violet!15}\textbf{Reflection:} Judgment = INSUFFICIENT\_USELESS \\
  & \cellcolor{violet!15}\quad ``The initial approach was too narrow... The core strategic error was not breaking the problem into smaller, more general steps.'' \\
  \midrule
  \textbf{Turn 2} & 
  \cellcolor{violet!15}\textbf{Recovery Planning:} \\
  & \cellcolor{violet!15}\quad Strategy Shift Reflection: ``Instead of starting with the player's name, we should first identify the season or year when these stats might have been achieved, or search for a list of Stanford receivers with notable seasons.'' \\
  & \cellcolor{violet!15}\quad Updated Analysis: ``Shift focus to general sources of Stanford football statistics or notable players.'' \\
  & \cellcolor{violet!15}\quad Next Queries: [``Stanford football all-time receiving leaders'', ``Stanford football players with 14 TDs in a season''] \\
  & \textbf{Exploration (Query: ``Stanford football all-time receiving leaders''):} \\
  & \quad \textit{KG:} Linked to ``Stanford Cardinal football statistical leaders'' (Q19878473) \\
  & \quad \quad Relation Discovery: Found 1 relation - ``sport'' (P641) $\rightarrow$ American football \\
  & \quad \quad Fact Pruning: Kept 0 facts (too general, not useful for identifying specific players) \\
  & \quad \textit{Wikipedia:} Found ``Stanford Cardinal football statistical leaders'' page with ``Receiving'' section 
  \\
  & \cellcolor{cyan!20}\quad \quad ``Receiving'' section containing 9 statistical tables, extracted the following information from table: \\
  & \cellcolor{cyan!20}\quad \quad \quad ``Single season 6'' (receiving yards): James Lofton - 1010 yards (1977) \\
  & \cellcolor{cyan!20}\quad \quad \quad ``Single season 7'' (receiving TDs): James Lofton - 14 TDs (1977); J.J. Arcega-Whiteside - 14 TDs (2018) \\
  & \cellcolor{cyan!20}\textbf{Key Discovery:} Identified ``James Lofton'' as candidate with 2/3 statistics matched (1010 yards + 14 TDs in 1977) \\
  & \textbf{Synthesis:} Strong candidate found but 57 receptions unconfirmed \\
  & \textbf{Reflection:} Judgment = INSUFFICIENT\_USEFUL - Continue with James Lofton \\
  \midrule
  \textbf{Turn 3} & 
  \textbf{Planning Phase:} \\
  & \quad Query: ``James Lofton 1977 season stats'' \\
  & \quad Entities: ``James Lofton'' \\
  & \textbf{Exploration Phase:} \\
  & \quad \textit{KG:} Linked to James Lofton (Q1680714) \\
  & \cellcolor{cyan!20}\quad \quad \textbf{Identity Check:} \texttt{educated at} $\rightarrow$ Stanford; \texttt{member of team} $\rightarrow$ Stanford Cardinal. \\
  & \cellcolor{cyan!20}\quad \quad \textbf{Attribute Retrieval:} Found alias ``\textbf{James David Lofton}'' (Directly reveals middle name) \\
  & \quad \textit{Wikipedia:} Found ``James Lofton'' page \\
  & \cellcolor{cyan!20}\quad \quad \textbf{From ``College career'' section:} Confirmed ``57 receptions for 1,
  010 yards and 14 touchdowns during the 1977 season'' \\
  & \cellcolor{violet!15}\textbf{Synthesis:} KG confirmed identity \& name; Text verified the missing statistical constraint. Middle name derivation, complete. \\
  & \cellcolor{violet!15}\textbf{Reflection:} Judgment = SUFFICIENT \\
  & \cellcolor{violet!15}\textbf{Final Answer:} ``The middle name of the former Stanford Cardinals player is David.'' \\
  & \cellcolor{violet!15}\textbf{Outcome:} \textbf{SUCCESS} in 3 turns \\
  \bottomrule
  \end{tabularx}
  \caption{Reasoning trace of CoG. Reflection and Proactive Planning diagnose the initial failure and recover from dead ends. Adaptive Entity Linking identifies implicitly useful entities (``1977 Stanford Cardinals football team'') not directly mentioned in the question. Finally, Text Entity Utilization effectively extracts the target entity (``James Lofton'') from text to guide the final answer derivation. Purple highlights show successful reflection and strategic adjustment; blue highlights show key information extraction from dual sources (KG structure + Wikipedia text).}
  \label{tab:case_cog}
\end{table*}

\section{Prompts}
\label{sec:prompts}

This section provides the main prompt used in CoG to guide LLM reasoning at each stage of the cognitive cycle.

Table~\ref{tab:prompt_init_plan} presents the prompt used for the \textbf{initial planning phase} (Section~\ref{sec:planning}). 
The prompt includes four examples to guide the LLM in handling diverse multi-hop reasoning scenarios.

Tables~\ref{tab:prompt_entity_link}--\ref{tab:prompt_fact_prune} present the prompts for the three-stage \textbf{KG Exploration} (Section~\ref{sec:exploration}): Entity Linking, Relation Discovery, and Fact Pruning, respectively.
These prompts demonstrate how CoG leverages in-context learning to achieve precise, schema-aware navigation over large-scale KGs.

Tables~\ref{tab:prompt_disambiguate}--\ref{tab:prompt_detail} detail the prompts used in the \textbf{Text Exploration module} (Section~\ref{sec:exploration}), covering Adaptive Page Selection (Tables~\ref{tab:prompt_disambiguate} and~\ref{tab:prompt_refine}) and Hierarchical Reading (Tables~\ref{tab:prompt_skimming} and~\ref{tab:prompt_detail}).

Table~\ref{tab:prompt_synthesis} presents the prompt for the \textbf{Synthesis module}. Tables~\ref{tab:example_kg_evidence} and \ref{tab:example_text_evidence} provide examples of structured evidence formats for KG and text, respectively, that populate the \texttt{\$\{evidence\_blocks\}} placeholder in the Synthesis prompt.

Tables~\ref{tab:prompt_cont_explore} and \ref{tab:prompt_recover} detail the prompts for the \textbf{Reflection module} (Section~\ref{sec:synthesis_reflection}), enabling the system to either deepen exploration based on new findings or pivot its strategy by diagnosing failures from the interaction history.

\begin{table*}[t]
  \centering
  \small
  \begin{tcolorbox}[
      enhanced,
      colback=blue!3,           
      colframe=blue!40!black,   
      boxrule=1pt,
      arc=2pt,
      left=5pt, right=5pt, top=5pt, bottom=5pt,
      fontupper=\small
  ]
  
\colorbox{blue!15}{\textbf{\strut ROLE}}\\[2pt]
You are a highly strategic retrieval planner. Your function is to analyze a complex question and identify the immediate, parallelizable search actions required to proceed.

\vspace{4pt}
\colorbox{cyan!15}{\textbf{\strut TASK}}\\[2pt]
Given a user's question, do NOT provide the final answer. Instead, determine the next logical search queries. A single question might require multiple parallel queries to gather the necessary initial information. Your plan must include your reasoning, a list of search queries, and a corresponding list of entities.

\vspace{4pt}
\colorbox{orange!20}{\textbf{\strut CONSTRAINTS}}\\[2pt]
- The number of queries in the `Query' list must not exceed 5.\\
- {\textbf{[CRITICAL]}} The `Query' and `Entities' lists must have the exact same number of items. For each query, the corresponding item in the `Entities' list must be the \textbf{single, most specific, and central named entity} being investigated in that query.\\
- Only output the query/queries for the immediate next step. Do not plan, describe, or allude to any subsequent steps.

\vspace{4pt}
\colorbox{green!15}{\textbf{\strut OUTPUT FORMAT}}\\[2pt]
Your output must follow this exact structure, with no additional commentary:\\[3pt]
\texttt{\textbf{Analysis:}} Your thought process explaining why these next steps are necessary and what information is being sought.\\
\texttt{\textbf{Query:}} [A Python-style list of concise and effective search query strings.]\\
\texttt{\textbf{Entities:}} [A Python-style list of the core entities in the Query, corresponding one-to-one with the Query list.]

\vspace{4pt}
\rule{\linewidth}{0.8pt}

\vspace{3pt}
\colorbox{purple!35}{\textbf{\strut EXAMPLES}}

\vspace{3pt}
\textit{\textbf{Example 1:}}\\
\textbf{Question:} What is the nationality of one of the men for whom the Schumann–Runge bands are named?\\[2pt]
\textbf{Output:}\\
\texttt{Analysis:} The question requires the nationality of a person. Before I can find their nationality, I must first identify the individuals the "Schumann–Runge bands" are named after.\\
\texttt{Query:} ["who are the Schumann-Runge bands named after"]\\
\texttt{Entities:} ["Schumann–Runge bands"]

\vspace{4pt}
\textit{... (Other examples omitted for brevity) ...}

\vspace{4pt}
\textit{Example 4:}\\
\textbf{Question:} Where did the leader of the largest European country after the collapse of the country that denied anything more than an advisory role in the Korean war die?\\
\textbf{Output:}\\
\texttt{Analysis:} The question is complex and requires multiple pieces of information. First, I need to identify which country denied anything more than an advisory role in the Korean War. [...] I will begin by identifying the country that denied a combat role in the Korean War.\\
\texttt{Query:} ["country that denied anything more than an advisory role in Korean War"]\\
\texttt{Entities:} ["Korean War"]

\vspace{3pt}
\rule{\linewidth}{0.8pt}

\vspace{3pt}
\colorbox{blue!15}{\textbf{\strut YOUR TASK}}\\[2pt]
\textbf{Question:} \texttt{\$\{question\}}\\[3pt]
\textbf{Output:}

  \end{tcolorbox}
  \caption{Prompt template for the Initial Planning module. The LLM is instructed to analyze the question, generate parallelizable sub-queries, and identify corresponding anchor entities. This structured output initializes the exploration cycle.}
  \label{tab:prompt_init_plan}
\end{table*}

\begin{table*}[t]
  \centering
  \small
  \begin{tcolorbox}[
      enhanced,
      colback=blue!3,
      colframe=blue!40!black,
      boxrule=1pt,
      arc=2pt,
      left=5pt, right=5pt, top=5pt, bottom=5pt,
      fontupper=\small
  ]
  
  \colorbox{blue!15}{\textbf{\strut ROLE}}\\[2pt]
  You are an expert in Knowledge Graph entity linking. Your task is to disambiguate an entity mention from a user's question by matching it to the correct entity in a knowledge graph, using the surrounding context.
  
  \vspace{4pt}
  \colorbox{purple!15}{\textbf{\strut CONTEXT}}\\[2pt]
  - \textbf{Original Question:} \texttt{\$\{question\}}\\
  - \textbf{Overall Plan (Analysis):} \texttt{\$\{analysis\}}\\
  - \textbf{Current Sub-Query:} \texttt{\$\{query\}}\\
  - \textbf{Entity Mention to Link:} \texttt{\$\{entity\}}
  
  \vspace{4pt}
  \colorbox{blue!15}{\textbf{\strut CANDIDATE ENTITIES}}\\[2pt]
  Here are the top candidate entities in the Knowledge Graph, sorted by a preliminary relevance score. Each candidate includes its Wikidata QID, label, description, aliases, popularity, some of its neighbors, and relevance score.\\[3pt]
  \texttt{\textcolor{orange}{\$\{formatted\_candidates\}}}
  
  \vspace{4pt}
  \colorbox{cyan!15}{\textbf{\strut YOUR TASK}}\\[2pt]
  Your task is to critically evaluate the candidate entities based on the provided CONTEXT. Your goal is to either identify the single correct entity OR determine that no suitable match exists.
  \begin{itemize}[leftmargin=*, nosep]
      \item \textbf{Analyze the Context:} Carefully review the 'Original Question' and 'Overall Plan'. What are the key details about the entity \texttt{\$\{entity\}} (e.g., their time period, relationships, role)?
      \item \textbf{Evaluate Each Candidate:} For each candidate, compare its \texttt{Description}, \texttt{Aliases}, and \texttt{Neighborhood} against the context. A correct match should be consistent with the context.
      \item \textbf{Make a Decision:}
      \begin{itemize}
          \item \textbf{If, and only if,} you find one candidate that is a confident and accurate match for \texttt{\$\{entity\}} based on all available information, your output should be its QID on a single line.
          \item \textbf{If none of the candidates are a confident match,} or if their key details contradict the context, you MUST output the single word: \texttt{NO\_MATCH}.
      \end{itemize}
  \end{itemize}
  
  \vspace{4pt}
  \colorbox{green!15}{\textbf{\strut OUTPUT FORMAT}}\\[2pt]
  \begin{itemize}[leftmargin=*, nosep]
      \item If a confident match is found, your entire output MUST be only the QID of that entity (e.g., Q12345).
      \item If no confident match is found, your entire output MUST be the single word: \texttt{NO\_MATCH}.
      \item Your output must not contain any other text, explanation, or reasoning.
  \end{itemize}

  \vspace{4pt}
  \rule{\linewidth}{0.8pt}
  
  \vspace{3pt}
  \colorbox{blue!15}{\textbf{\strut YOUR RESPONSE:}}
  
  \end{tcolorbox}
  \caption{Prompt template for the Entity Linking module. The LLM is tasked with precise disambiguation by comparing candidate profiles against the question context. The placeholder \texttt{\textcolor{orange}{\$\{formatted\_candidates\}}} is populated with structured candidate data, as illustrated in Table~\ref{tab:example_entity_link}.}
  \label{tab:prompt_entity_link}
\end{table*}

\begin{table*}[t]
  \centering
  \small
  \begin{tcolorbox}[
    enhanced,
    colback=teal!3,
    colframe=teal!60!black!40,   
    boxrule=0.8pt,            
    arc=2pt,
    left=5pt, right=5pt, top=5pt, bottom=5pt,
    fontupper=\small\ttfamily, 
    title=\textbf{Example of Formatted Candidates},
    coltitle=black,           
    colbacktitle=teal!60!gray!30,     
    attach boxed title to top left={yshift=-2mm, xshift=2mm}, 
    boxed title style={boxrule=0.5pt, colframe=gray!50}
  ]
  \texttt{---}\\
  \texttt{[Entity]}\\
  \texttt{\ \ - QID: Q716680}\\
  \texttt{\ \ - Label: Steve McQueen}\\
  \texttt{\ \ - Description: American actor (1930-1980)}\\
  \texttt{\ \ - Aliases: Terrence Stephen McQueen, King of Cool}\\
  \texttt{\ \ - Popularity (Degree): In: 154, Out: 45, Attr: 22}\\
  \texttt{[Neighborhood]}\\
  \texttt{\ \ - Outgoing Relations (as Head):}\\
  \texttt{\ \ \ \ - spouse: Neile Adams, Ali MacGraw, Barbara Minty}\\
  \texttt{\ \ \ \ - child: Chad McQueen, Terry Leslie McQueen}\\
  \texttt{[Scores]}\\
  \texttt{\ \ - Final: 0.952 (Name similarity=1.000, Popularity=0.850, Description=0.920, Neighbor=0.880)}\\
  \texttt{---}\\
  \textit{... (Other candidates omitted) ...}
  \end{tcolorbox}
  \caption{An example of the formatted candidate data populated into the \texttt{\textcolor{orange}{\$\{formatted\_candidates\}}} slot in Table~\ref{tab:prompt_entity_link}. This structured format provides the LLM with comprehensive entity details (profile, neighborhood, and retrieval scores) for disambiguation.}
  \label{tab:example_entity_link}
\end{table*}

\begin{table*}[t]
  \centering
  \small
  \begin{tcolorbox}[
      enhanced,
      colback=blue!3,
      colframe=blue!40!black,
      boxrule=1pt,
      arc=2pt,
      left=5pt, right=5pt, top=5pt, bottom=5pt,
      fontupper=\small
  ]
  
  \colorbox{blue!15}{\textbf{\strut ROLE}}\\[2pt]
  You are an expert Knowledge Graph analyzer. Your task is to select the most relevant relations (properties) of a given entity that will help answer a specific question.
  
  \vspace{4pt}
  \colorbox{purple!15}{\textbf{\strut CONTEXT}}\\[2pt]
  - \textbf{Original Question:} \texttt{\$\{question\}}\\
  - \textbf{Overall Plan (Analysis):} \texttt{\$\{analysis\}}\\
  - \textbf{Current Sub-Query:} \texttt{\$\{query\}}\\
  - \textbf{Entity in Focus:} \texttt{\$\{entity\_label\}} (\texttt{\$\{entity\_qid\}}): \texttt{\$\{entity\_description\}}
  
  \vspace{4pt}
  \colorbox{blue!15}{\textbf{\strut AVAILABLE RELATIONS}}\\[2pt]
  The entity \texttt{\$\{entity\_label\}} is connected to \texttt{\$\{total\_relations\_count\}} relations (Outgoing: \texttt{\$\{outgoing\_count\}}, Incoming: \texttt{\$\{incoming\_count\}}) in the Knowledge Graph. Below is a list of all relations connected to the entity. Each relation includes its label, PID, and a frequency indicating how many other entities in the graph are linked via this relation.\\[3pt]
  \textbf{Outgoing Relations:}\\
  \texttt{\$\{outgoing\_relations\}}\\[3pt]
  \textbf{Incoming Relations:}\\
  \texttt{\$\{incoming\_relations\}}
  
  \vspace{4pt}
  \colorbox{cyan!15}{\textbf{\strut YOUR TASK}}\\[2pt]
  Based on the \textbf{CONTEXT}, select relations that will help find the answer using a two-pronged approach:
  \begin{enumerate}[leftmargin=*, nosep]
      \item \textbf{Direct Relations}: Select relations that seem to directly contain the answer.
      \item \textbf{Exploratory Relations}: If no direct relations exist, select relations that could lead to intermediate entities, which might then contain the answer.
  \end{enumerate}
  
  \vspace{4pt}
  \colorbox{green!15}{\textbf{\strut OUTPUT FORMAT}}\\[2pt]
  Your output must follow this exact structure, with no additional commentary:\\[3pt]
  \texttt{\textbf{Reasoning:}} Your brief thought process explaining why you chose these specific relations based on the question and the entity.\\
  \texttt{\textbf{Selected PIDs:}} [A Python-style list of strings, where each string is the PID of a selected relation (e.g., ["P31", "P17"]). If no relations seem relevant, output []]
  
  \vspace{4pt}
  \rule{\linewidth}{0.8pt}
  
  \vspace{3pt}
  \colorbox{blue!15}{\textbf{\strut YOUR RESPONSE:}}

  \end{tcolorbox}
  \caption{Prompt template for the Relation Discovery module. The LLM selects a subset of relations relevant to the sub-query to prune the search space at the schema level.}
  \label{tab:prompt_relation_discovery}
\end{table*}

\begin{table*}[t]
  \centering
  \small
  \begin{tcolorbox}[
      enhanced,
      colback=blue!3,
      colframe=blue!40!black,
      boxrule=1pt,
      arc=2pt,
      left=5pt, right=5pt, top=5pt, bottom=5pt,
      fontupper=\small
  ]
  
  \colorbox{blue!15}{\textbf{\strut ROLE}}\\[2pt]
  You are an expert assistant specializing in knowledge graph analysis for question answering. Your mission is to intelligently prune a list of facts, guiding a multi-step exploration process.
  
  \vspace{4pt}
  \colorbox{purple!15}{\textbf{\strut CONTEXT}}\\[2pt]
  - \textbf{Original Question:} \texttt{\$\{question\}}\\
  - \textbf{Overall Plan (Analysis):} \texttt{\$\{analysis\}}\\
  - \textbf{Current Sub-Query:} \texttt{\$\{query\}}\\
  - \textbf{Entity in Focus:} \texttt{\$\{entity\_label\}}
  
  \vspace{4pt}
  \colorbox{blue!15}{\textbf{\strut AVAILABLE FACTS}}\\[2pt]
  Below is a structured list of all facts retrieved from the Knowledge Graph for the entity \texttt{\$\{entity\_label\}}. The facts are categorized by relation type (Outgoing or Incoming).\\[3pt]
  \texttt{\$\{formatted\_facts\}}
  
  \vspace{4pt}
  \colorbox{cyan!15}{\textbf{\strut YOUR TASK}}\\[2pt]
  Your goal is to decide which facts to keep. Evaluate the facts using a two-pronged approach:
  \begin{enumerate}[leftmargin=*, nosep]
      \item \textbf{Directly Relevant Facts}: Keep any facts that directly help answer the \textbf{Current Sub-Query}.
      \item \textbf{Promising Intermediate Entities}: Keep facts that represent entities which are promising stepping stones. A fact is "promising" if exploring it further is highly likely to lead to the answer or provide crucial context.
  \end{enumerate}
  Discard facts that are clearly irrelevant, noisy, or are just general information.
  
  \vspace{4pt}
  \colorbox{green!15}{\textbf{\strut OUTPUT FORMAT}}\\[2pt]
  Your output \textbf{MUST} be a single, valid JSON object with the following structure:
\begin{verbatim}
  {
    "reasoning": "Structured thought process...",
    "pruned_facts": {
      "outgoing": { "P17": ["kept fact labels"], ... },
      "incoming": { "P802": ["kept fact labels"], ... }
    }
  }
\end{verbatim}
  
  \vspace{4pt}
  \rule{\linewidth}{0.8pt}
  
  \vspace{3pt}
  \colorbox{blue!15}{\textbf{\strut YOUR RESPONSE:}}

  \end{tcolorbox}
  \caption{Prompt template for the Fact Pruning module. The LLM performs fine-grained filtering of facts to retain only critical evidence or promising exploration leads.}
  \label{tab:prompt_fact_prune}
\end{table*}

\begin{table*}[t]
  \centering
  \small
  \begin{tcolorbox}[
      enhanced,
      colback=blue!3,
      colframe=blue!40!black,
      boxrule=1pt,
      arc=2pt,
      left=5pt, right=5pt, top=5pt, bottom=5pt,
      fontupper=\small
  ]
  
  \colorbox{blue!15}{\textbf{\strut ROLE}}\\[2pt]
  You are a research agent resolving an ambiguous search query on Wikipedia.
  
  \vspace{4pt}
  \colorbox{cyan!15}{\textbf{\strut TASK}}\\[2pt]
  Your search for "\texttt{\$\{entity\}}" led to a disambiguation page with multiple possible meanings. You must carefully analyze the full context provided below to choose the single most relevant link from the options.
  
  \vspace{4pt}
  \colorbox{purple!15}{\textbf{\strut CONTEXT}}\\[2pt]
  - \textbf{Original Question:} \texttt{\$\{question\}}\\
  - \textbf{Overall Plan (Analysis):} \texttt{\$\{analysis\}}\\
  - \textbf{Current Sub-Task (Query):} \texttt{\$\{query\}}\\
  - \textbf{Ambiguous Search Query:} "\texttt{\$\{entity\}}"
  
  \vspace{4pt}
  \colorbox{blue!15}{\textbf{\strut OPTIONS}}\\[2pt]
  \texttt{\$\{options\_list\}}
  
  \vspace{4pt}
  \colorbox{green!15}{\textbf{\strut YOUR DECISION}}\\[2pt]
  Analyze the options based on the context to select the best fit. Your output must be a single line containing only the exact title of the chosen page from the list. If none of the options seem relevant for answering the original question, output the string "NO\_MATCH".
  
  \vspace{4pt}
  \rule{\linewidth}{0.8pt}
  
  \vspace{3pt}
  \colorbox{blue!15}{\textbf{\strut YOUR RESPONSE:}}
  
  \end{tcolorbox}
  \caption{Prompt template for the \textbf{Disambiguation} step in Adaptive Page Selection. The LLM resolves ambiguous entities by selecting the correct page from disambiguation options based on context.}
  \label{tab:prompt_disambiguate}
\end{table*}

\begin{table*}[t]
  \centering
  \small
  \begin{tcolorbox}[
      enhanced,
      colback=blue!3,
      colframe=blue!40!black,
      boxrule=1pt,
      arc=2pt,
      left=5pt, right=5pt, top=5pt, bottom=5pt,
      fontupper=\small
  ]
  
  \colorbox{blue!15}{\textbf{\strut ROLE}}\\[2pt]
  You are a research agent trying to find the right information on Wikipedia after a search query failed.
  
  \vspace{4pt}
  \colorbox{cyan!15}{\textbf{\strut TASK}}\\[2pt]
  Your search for "\texttt{\$\{entity\}}" did not find a matching Wikipedia page. Use the contextual information and the suggestions provided to decide on the best query for your next attempt.
  
  \vspace{4pt}
  \colorbox{purple!15}{\textbf{\strut CONTEXT}}\\[2pt]
  - \textbf{Original Question:} \texttt{\$\{question\}}\\
  - \textbf{Overall Plan (Analysis):} \texttt{\$\{analysis\}}\\
  - \textbf{Current Sub-Task (Query):} \texttt{\$\{query\}}\\
  - \textbf{Failed Search Query:} "\texttt{\$\{entity\}}"
  
  \vspace{4pt}
  \colorbox{blue!15}{\textbf{\strut SUGGESTIONS}}\\[2pt]
  Below are suggestions from the Wikipedia search API to help you refine your query.\\[3pt]
  \textbf{"Did you mean?":} This is the API's top recommendation, often correcting a typo or suggesting a more standard page title.\\
  - Suggestion: \texttt{\$\{suggestion\}}\\[3pt]
  \textbf{Similar Pages Found:} These are pages with titles that are textually similar to your failed query. One of them might be the correct page under a slightly different name.\\
  - Similar Pages: \texttt{\$\{search\_results\_list\}}
  
  \vspace{4pt}
  \colorbox{green!15}{\textbf{\strut YOUR DECISION}}\\[2pt]
  Based on the context and the suggestions, choose the best query for your next search attempt. This could be the "Did you mean?" suggestion, one of the similar pages, or a completely new query you formulate based on the feedback from this failed search. Your output must be a single line containing only the new search query. If none of the suggestions seem useful for answering the original question, output the string "NO\_MATCH".
  
  \vspace{4pt}
  \rule{\linewidth}{0.8pt}
  
  \vspace{3pt}
  \colorbox{blue!15}{\textbf{\strut YOUR RESPONSE:}}
  
  \end{tcolorbox}
  \caption{Prompt template for the \textbf{Search Refinement} step in Adaptive Page Selection. The LLM uses API feedback (corrections and similar pages) to recover from search failures.}
  \label{tab:prompt_refine}
\end{table*}

\begin{table*}[t]
  \centering
  \small
  \begin{tcolorbox}[
      enhanced,
      colback=blue!3,
      colframe=blue!40!black,
      boxrule=1pt,
      arc=2pt,
      left=5pt, right=5pt, top=5pt, bottom=5pt,
      fontupper=\small
  ]
  
  \colorbox{blue!15}{\textbf{\strut ROLE}}\\[2pt]
  You are a research agent tasked with answering a complex question by navigating Wikipedia.
  
  \vspace{4pt}
  \colorbox{cyan!15}{\textbf{\strut TASK}}\\[2pt]
  You have been provided with the summary, an infobox table, and the section list of a Wikipedia page relevant to your current query. Your goal is to extract all useful information from the summary and infobox, identify promising sections for deeper investigation, or determine that the page is irrelevant.
  
  \vspace{4pt}
  \colorbox{purple!15}{\textbf{\strut CONTEXT}}\\[2pt]
  - \textbf{Original Question:} \texttt{\$\{question\}}\\
  - \textbf{Overall Plan (Analysis):} \texttt{\$\{analysis\}}\\
  - \textbf{Current Search Query:} "\texttt{\$\{query\}}"\\
  - \textbf{Retrieved Wikipedia Page Title:} "\texttt{\$\{page\_title\}}"
  
  \vspace{4pt}
  \colorbox{blue!15}{\textbf{\strut AVAILABLE INFORMATION}}\\[2pt]
  \textbf{Page Summary:}\\
  \texttt{\$\{page\_summary\}}\\[3pt]
  \textbf{Summary Infobox:}\\
  \texttt{\$\{infobox\_table\}}\\[3pt]
  \textbf{Page Sections (Top-Level):}\\
  \texttt{\$\{page\_sections\}}
  
  \vspace{4pt}
  \colorbox{green!15}{\textbf{\strut YOUR DECISION}}\\[2pt]
  Based on the information above, what is your next best action? Choose one of the following two options.
  \begin{enumerate}[leftmargin=*, nosep]
      \item \textbf{If the page seems relevant:}
      \begin{itemize}[nosep]
          \item Extract \textbf{Direct Information} and \textbf{Promising Clues} from the summary/infobox.
          \item Identify sections for deeper investigation.
      \end{itemize}
      \item \textbf{If the page is clearly irrelevant:} Explain why and suggest a better search query.
  \end{enumerate}
  
  \vspace{4pt}
  \rule{\linewidth}{0.8pt}
  
  \vspace{3pt}
  \colorbox{blue!15}{\textbf{\strut YOUR RESPONSE:}}
  
  \end{tcolorbox}
  \caption{Prompt template for the Global Skimming module. The LLM assesses page relevance, extracts high-level evidence from the summary and infobox, and selects promising sections for detailed reading.}
  \label{tab:prompt_skimming}
\end{table*}

\begin{table*}[t]
  \centering
  \small
  \begin{tcolorbox}[
      enhanced,
      colback=blue!3,
      colframe=blue!40!black,
      boxrule=1pt,
      arc=2pt,
      left=5pt, right=5pt, top=5pt, bottom=5pt,
      fontupper=\small
  ]
  
  \colorbox{blue!15}{\textbf{\strut ROLE}}\\[2pt]
  You are a meticulous research agent. Your task is to analyze a section of a Wikipedia page that contains both text and tables to extract relevant information for a given query.
  
  \vspace{4pt}
  \colorbox{cyan!15}{\textbf{\strut TASK}}\\[2pt]
  You have been given several retrieved text chunks and a preview of all tables found within a specific Wikipedia section. Your goal is to:
  \begin{enumerate}[leftmargin=*, nosep]
      \item Extract \textbf{Direct Information} and \textbf{Promising Clues} from BOTH the \textbf{Retrieved Text Chunks} and the \textbf{Table Previews}.
      \item Analyze table previews. If a preview is insufficient but promising, select the table for a full read.
      \item Provide a single, unified rationale that explains both your information extraction findings and your table selections.
  \end{enumerate}
  
  \vspace{4pt}
  \colorbox{purple!15}{\textbf{\strut CONTEXT}}\\[2pt]
  - \textbf{Original Question:} \texttt{\$\{question\}}\\
  - \textbf{Current Search Query:} "\texttt{\$\{query\}}"\\
  - \textbf{Section Being Investigated:} "\texttt{\$\{section\_title\}}"\\
  - \textbf{Section Exploration Rationale (during summary reading):} \texttt{\$\{exploration\_rationale\}}
  
  \vspace{4pt}
  \colorbox{blue!15}{\textbf{\strut AVAILABLE INFORMATION}}\\[2pt]
  \textbf{1. Retrieved Text Chunks:} (Top \texttt{\$\{k\}} relevant chunks)\\
  \texttt{\$\{context\_chunks\}}\\[3pt]
  \textbf{2. Table Previews:} (Previews of tables: \texttt{\$\{table\_names\_in\_section\}})\\
  \texttt{\$\{tables\_preview\}}
  
  \vspace{4pt}
  \colorbox{green!15}{\textbf{\strut OUTPUT FORMAT}}\\[2pt]
  \textbf{Rationale:} A consolidated explanation covering both text and tables.\\
  \textbf{Extracted Info:} Key info from text chunks and table previews. State "None" if nothing relevant is found.\\
  \textbf{Selected Tables:} List of table names requiring a full read (e.g., ["Awards"]). Empty list [] if none.
  
  \vspace{4pt}
  \rule{\linewidth}{0.8pt}
  
  \vspace{3pt}
  \colorbox{blue!15}{\textbf{\strut YOUR RESPONSE:}}
  
  \end{tcolorbox}
  \caption{Prompt template for the \textbf{Detailed Reading} phase. The LLM performs joint analysis of retrieved text chunks and table previews to synthesize fine-grained evidence.}
  \label{tab:prompt_detail}
\end{table*}

\begin{table*}[t]
  \centering
  \small
  \begin{tcolorbox}[
      enhanced,
      colback=blue!3,
      colframe=blue!40!black,
      boxrule=1pt,
      arc=2pt,
      left=5pt, right=5pt, top=5pt, bottom=5pt,
      fontupper=\small
  ]
  
  \colorbox{blue!15}{\textbf{\strut ROLE}}\\[2pt]
  You are a master AI strategist leading a multi-hop question-answering mission. Your task is to synthesize retrieved information from various sources, evaluate progress against the overall plan, and decide the most logical next step.
  
  \vspace{4pt}
  \colorbox{purple!15}{\textbf{\strut CONTEXT}}\\[2pt]
  - \textbf{Original Question:} \texttt{\$\{question\}}\\
  - \textbf{Overall Plan (Analysis):} \texttt{\$\{analysis\}}\\
  - \textbf{Notebook (Summary of Known Facts):}\\
  \texttt{\$\{notebook\}}\\
  - \textbf{Current Sub-Queries:} \texttt{\$\{queries\}}\\
  - \textbf{Core Entities in Sub-Queries:} \texttt{\$\{entities\}}
  
  \vspace{4pt}
  \colorbox{blue!15}{\textbf{\strut EVIDENCE}}\\[2pt]
  This section contains the information retrieved from different sources for all sub-queries executed in this turn.\\
  \texttt{\textcolor{orange}{\$\{evidence\_blocks\}}}
  
  \vspace{4pt}
  \colorbox{cyan!15}{\textbf{\strut YOUR TASK}}\\[2pt]
  Carefully review all evidence, and in conjunction with the \textbf{Original Question} and your \textbf{Overall Plan}, complete the following three steps:
  \begin{enumerate}[leftmargin=*, nosep]
      \item \textbf{Verbatim Information Extraction:} Meticulously extract all useful information. Your goal is to faithfully transfer potentially relevant information to your `Extracted Content'. Include:
      \begin{itemize}[nosep]
          \item \textbf{Direct Facts:} Core facts that directly contribute to answering the original question.
          \item \textbf{Promising Leads:} New entities or critical factual clues essential for guiding the next step.
          \item \textbf{[CRITICAL]} Do NOT summarize or rephrase. This is for extraction only.
      \end{itemize}
      \item \textbf{Think Step-by-Step:} Document your thought process.
      \begin{itemize}[nosep]
          \item Explain how the direct facts help answer the question.
          \item Discuss the potential value of promising leads and how they might be explored.
          \item Identify what key information is still missing.
          \item If evidence is useless, explain \textit{why} the current sub-queries failed.
      \end{itemize}
      \item \textbf{Make a Judgment:} Choose one of the following options:
      \begin{itemize}[nosep]
          \item \texttt{SUFFICIENT}: Information is adequate to generate a final, complete answer.
          \item \texttt{INSUFFICIENT\_USEFUL}: Valuable clues found, but more information is needed. Continue investigation based on the current findings.
          \item \texttt{INSUFFICIENT\_USELESS}: Information is irrelevant or has led to a dead end. A new strategy is needed.
      \end{itemize}
  \end{enumerate}
  
  \vspace{4pt}
  \colorbox{green!15}{\textbf{\strut OUTPUT FORMAT}}\\[2pt]
  \texttt{\textbf{Thought Process:}} Your step-by-step thinking process...\\
  \texttt{\textbf{Extracted Content:}} A structured collection of key facts and promising new leads...\\
  \texttt{\textbf{Judgment:}} SUFFICIENT, INSUFFICIENT\_USEFUL, or INSUFFICIENT\_USELESS.
  
  \vspace{4pt}
  \rule{\linewidth}{0.8pt}
  
  \vspace{3pt}
  \colorbox{blue!15}{\textbf{\strut YOUR RESPONSE:}}
  
  \end{tcolorbox}
  \caption{Prompt template for the \textbf{Synthesis} module. The LLM aggregates dual-source evidence, extracts key insights, and judges the current progress to determine the next action (answer, continue, or recover). Examples of the content for \texttt{\textcolor{orange}{\$\{evidence\_blocks\}}} are shown in Tables~\ref{tab:example_kg_evidence} and~\ref{tab:example_text_evidence}.}
  \label{tab:prompt_synthesis}
\end{table*}

\begin{table*}[h]
  \centering
  \small
  \begin{tcolorbox}[
    enhanced,
    colback=teal!3,
    colframe=teal!60!black!40,   
    boxrule=0.8pt,            
    arc=2pt,
    left=5pt, right=5pt, top=5pt, bottom=5pt,
    fontupper=\small\ttfamily, 
    title=\textbf{Example of Formatted Candidates},
    coltitle=black,           
    colbacktitle=teal!60!gray!30,     
    attach boxed title to top left={yshift=-2mm, xshift=2mm}, 
    boxed title style={boxrule=0.5pt, colframe=gray!50}
  ]
  \texttt{KG Exploration for Mention: "John Cage"}\\
  \\
  \texttt{[Entity Linking]}\\
  \texttt{- Linked Entity: John Cage (Q180727)}\\
  \texttt{- Description: American composer and music theorist}\\
  \texttt{- Candidates (top-4 preview):}\\
  \texttt{\ \ - John Cage (Q180727): American composer and music theorist}\\
  \texttt{\ \ - John Cage (Q5347597): Fictional character from Ally McBeal}\\
  \texttt{- Linking Rationale: The query context discusses "4'33" and "prepared piano", which are signature works of the composer John Cage.}\\
  \\
  \texttt{[Retrieved Facts for "John Cage"]}\\
  \texttt{[Attributes]}\\
  \texttt{- date of birth: ['1912-09-05']}\\
  \texttt{- occupation: ['composer', 'philosopher', 'artist']}\\
  \texttt{[Outgoing]}\\
  \texttt{- notable work: ["4'33''", 'Music of Changes', 'Imaginary Landscape No. 4']}\\
  \texttt{- student of: ['Arnold Schoenberg', 'Henry Cowell']}\\
  \texttt{[Incoming]}\\
  \texttt{- influenced by: ['Erik Satie', 'Marcel Duchamp', 'D. T. Suzuki']}\\
  \\
  \texttt{[Filtering Summary]}\\
  \texttt{- Relations explored: 5 of 21 total}\\
  \texttt{- Facts kept: 12 of 43 total}\\
  \\
  \texttt{[Reasoning]}\\
  \texttt{- Relation Selection: Selected relations regarding works, influences, and personal life relevant to his artistic development.}\\
  \texttt{- Fact Pruning: Kept only major works and key figures like Schoenberg and Cunningham; discarded minor administrative categories.}
  \end{tcolorbox}
  \caption{An example of structured KG evidence formatted for the \texttt{\textcolor{orange}{\$\{evidence\_blocks\}}} in the Synthesis module. It includes the linked entity, verified facts (attributes and relations), and a reasoning trace for the pruning process.}
  \label{tab:example_kg_evidence}
\end{table*}

\begin{table*}[h]
  \centering
  \small
  \begin{tcolorbox}[
    enhanced,
    colback=teal!3,
    colframe=teal!60!black!40,   
    boxrule=0.8pt,            
    arc=2pt,
    left=5pt, right=5pt, top=5pt, bottom=5pt,
    fontupper=\small\ttfamily, 
    title=\textbf{Example of Formatted Candidates},
    coltitle=black,           
    colbacktitle=teal!60!gray!30,     
    attach boxed title to top left={yshift=-2mm, xshift=2mm}, 
    boxed title style={boxrule=0.5pt, colframe=gray!50}
  ]
  \ttfamily
  Wikipedia Retrieval Results for entity: "John Cage"
  \par\vspace{0.5em}
  \textbf{Information Extraction:}
  \begin{itemize}[leftmargin=*, label=-, nosep]
      \item \textbf{Page Title:} John Cage
      \item \textbf{All Page Sections:} ['Life', 'Music', 'Visual art...', ...]
  \end{itemize}
  \vspace{0.5em}
  \begin{itemize}[leftmargin=*, label=-, nosep]
      \item \textbf{A. Skimming Summary and Section Selection:}
      \begin{itemize}[leftmargin=*, label=-, nosep]
          \item \textbf{Rationale:} The summary highlights Cage as a pioneer of indeterminacy and the prepared piano. To understand his specific methods, we need to explore the "Music" section.
          \item \textbf{Selected Sections for Deeper Analysis:} ['Music']
      \end{itemize}
  \end{itemize}
  \vspace{0.5em}
  \begin{itemize}[leftmargin=*, label=-, nosep]
      \item \textbf{B. Extracted Content:}
      \begin{itemize}[leftmargin=*, label=-, nosep]
          \item \textbf{From Summary:}\\
          John Cage was a pioneer of indeterminacy in music. Best known for "4'33''", a silent composition. He developed the "prepared piano" and used the "I Ching" as a standard composition tool.
          \item \textbf{From Sections:}
          \begin{itemize}[leftmargin=*, label=-, nosep]
              \item \textbf{Section: "Music"}
              \begin{itemize}[leftmargin=*, label=-, nosep]
                  \item \textbf{Rationale for section processing:} Contains detailed descriptions of his composition techniques like chance operations and rhythmic structures.
                  \item \textbf{Extracted from Text:}\\
                  \hspace*{1em}- In 1951, Cage started using the \textbf{I Ching} to compose using chance, imitating nature's manner of operation.\\
                  \hspace*{1em}- \textbf{Music of Changes} (1951) was the first major work created using this method.\\
                  \hspace*{1em}- \textbf{Cheap Imitation} (1969) is a chance-controlled reworking of Erik Satie's Socrate.
              \end{itemize}
          \end{itemize}
      \end{itemize}
  \end{itemize}
  \end{tcolorbox}
  \caption{An example of Text evidence formatted for the \texttt{\textcolor{orange}{\$\{evidence\_blocks\}}} in the Synthesis module. It summarizes the high-level page content and provides fine-grained details extracted from specific sections relevant to the query.}
  \label{tab:example_text_evidence}
\end{table*}

\begin{table*}[t]
  \centering
  \small
  \begin{tcolorbox}[
      enhanced,
      colback=blue!3,
      colframe=blue!40!black,
      boxrule=1pt,
      arc=2pt,
      left=5pt, right=5pt, top=5pt, bottom=5pt,
      fontupper=\small
  ]
  
  \colorbox{blue!15}{\textbf{\strut ROLE}}\\[2pt]
  You are a master AI strategist leading a multi-hop question-answering mission. Your task is to plan the next step of the investigation after an information-gathering turn that was useful but insufficient.
  
  \vspace{4pt}
  \colorbox{purple!15}{\textbf{\strut CONTEXT}}\\[2pt]
  - \textbf{Original Question:} \texttt{\$\{question\}}\\
  - \textbf{Current Notebook:} \texttt{\$\{notebook\}}\\
  - \textbf{Previous Overall Plan:} \texttt{\$\{analysis\}}\\
  - \textbf{Previous Sub-Queries:} \texttt{\$\{queries\}}\\
  - \textbf{Candidate Entities Pool:} \texttt{\$\{candidate\_entities\_pool\}}
  
  \vspace{4pt}
  \colorbox{blue!15}{\textbf{\strut NEW FINDINGS}}\\[2pt]
  - \textbf{Thought Process:} \texttt{\$\{thought\_process\}}\\
  - \textbf{Extracted Content:} \texttt{\$\{extracted\_content\}}
  
  \vspace{4pt}
  \colorbox{cyan!15}{\textbf{\strut YOUR TASK}}\\[2pt]
  Plan the next round of investigation to dig deeper and bridge information gaps.
  \begin{enumerate}[leftmargin=*, nosep]
      \item \textbf{Update Notebook:} Combine current notebook with new findings into a single, coherent summary, retaining all unique and relevant details.
      \item \textbf{Update Analysis:} Revise the overall plan to reflect new understanding. Explain what the immediate next step should focus on and why.
      \item \textbf{Plan Next Queries:} Define a new set of sub-queries and corresponding core entities based on the updated analysis.
      \item \textbf{Manage Candidate Pool:} Review existing leads and identify new promising entities from the new findings. Update the pool by adding new leads and removing promoted ones.
  \end{enumerate}
  
  \vspace{4pt}
  \colorbox{green!15}{\textbf{\strut OUTPUT FORMAT}}\\[2pt]
  Your output must follow this exact structure:\\[3pt]
  \texttt{\textbf{Thought Process:}} Your step-by-step reasoning.\\
  \texttt{\textbf{Updated Notebook:}} The comprehensive summary of all known facts.\\
  \texttt{\textbf{Updated Analysis:}} Your revised analysis and plan.\\
  \texttt{\textbf{Next Queries:}} [List of concise and effective search query.]\\
  \texttt{\textbf{Next Entities:}} [List of core entities corresponding to queries]\\
  \texttt{\textbf{Updated Candidate Pool:}} [List of dictionaries, e.g., \{"entity": "Name", "reason": "..."\}]
  
  \vspace{4pt}
  \rule{\linewidth}{0.8pt}
  
  \vspace{3pt}
  \colorbox{blue!15}{\textbf{\strut YOUR RESPONSE:}}
  
  \end{tcolorbox}
  \caption{Prompt template for \textbf{Continue Exploration} in the Reflection module. The LLM updates the notebook and analysis based on new findings and plans the next steps to deepen the investigation.}
  \label{tab:prompt_cont_explore}
\end{table*}

\begin{table*}[t]
  \centering
  \small
  \begin{tcolorbox}[
      enhanced,
      colback=blue!3,
      colframe=blue!40!black,
      boxrule=1pt,
      arc=2pt,
      left=5pt, right=5pt, top=5pt, bottom=5pt,
      fontupper=\small
  ]
  
  \colorbox{blue!15}{\textbf{\strut ROLE}}\\[2pt]
  You are a master AI strategist leading a multi-hop question-answering mission. Your task is to recover from a failed information-gathering turn where the retrieved evidence was useless.
  
  \vspace{4pt}
  \colorbox{purple!15}{\textbf{\strut CONTEXT}}\\[2pt]
  - \textbf{Original Question:} \texttt{\$\{question\}}\\
  - \textbf{Current Notebook:} \texttt{\$\{notebook\}}\\
  - \textbf{Candidate Entities Pool:} \texttt{\$\{candidate\_entities\_pool\}}\\
  - \textbf{Interaction History:} \texttt{\$\{interaction\_history\}}
  
  \vspace{4pt}
  \colorbox{red!30}{\textbf{\strut FAILED TURN DETAILS}}\\[2pt]
  The last exploration turn was deemed unproductive. Here's what went wrong:\\
  - \textbf{Previous Overall Plan:} \texttt{\$\{analysis\}}\\
  - \textbf{Failed Sub-Queries:} \texttt{\$\{queries\}}\\
  - \textbf{Failed Core Entities:} \texttt{\$\{entities\}}\\
  - \textbf{Extracted Content:} \texttt{\$\{extracted\_content\}}\\
  - \textbf{Reasoning for Failure:} \texttt{\$\{thought\_process\}}
  
  \vspace{4pt}
  \colorbox{cyan!15}{\textbf{\strut YOUR TASK}}\\[2pt]
  The previous approach hit a dead end. Critically reflect on the history, diagnose the error, and pivot the strategy.
  \begin{enumerate}[leftmargin=*, nosep]
      \item \textbf{Critical Reflection:} Analyze why the previous approach failed (e.g., flawed plan, wrong entities). Summarize the core strategic error and derive actionable \textbf{guiding principles} for the next attempt.
      \item \textbf{Update Analysis:} Propose a fundamentally new plan that leverages insights from reflection. Consider promoting leads from the Candidate Pool, re-examining the original question for missed keywords, or formulating queries from an entirely different angle.
      \item \textbf{Plan Next Queries:} Define a new set of queries that represent a clear change in direction from the failed ones.
      \item \textbf{Manage Candidate Pool:} Review the pool. Decide if existing leads are now high-priority, add any new promising leads discovered incidentally, and generate the updated pool by carrying over unused leads while removing promoted ones.
  \end{enumerate}
  
  \vspace{4pt}
  \colorbox{green!15}{\textbf{\strut OUTPUT FORMAT}}\\[2pt]
  Your output must follow this exact structure:\\[3pt]
  \texttt{\textbf{Thought Process:}} Critical reflection on failure and reasoning for the new plan.\\
  \texttt{\textbf{Updated Analysis:}} The revised analysis and new strategic direction.\\
  \texttt{\textbf{Next Queries:}} [List of concise and effective search query.]\\
  \texttt{\textbf{Next Entities:}} [List of core entities corresponding to queries.]\\
  \texttt{\textbf{Updated Candidate Pool:}} [List of updated candidate dictionaries, e.g., \{"entity": "Name", "reason": "..."\}]
  
  \vspace{4pt}
  \rule{\linewidth}{0.8pt}
  
  \vspace{3pt}
  \colorbox{blue!15}{\textbf{\strut YOUR RESPONSE:}}
  
  \end{tcolorbox}
  \caption{Prompt template for \textbf{Strategy Adjustment} in the Reflection module. The LLM diagnoses the failure using interaction history and pivots the strategy by generating entirely new queries or backtracking to candidate entities.}
  \label{tab:prompt_recover}
\end{table*}

\end{document}